\documentclass[letterpaper]{article} 
\usepackage{aaai21}  
\usepackage{times}  
\usepackage{helvet} 
\usepackage{courier}  
\usepackage[hyphens]{url}  
\usepackage{graphicx} 
\usepackage{amsmath,amssymb}
\usepackage{natbib}  
\usepackage{caption} 
\title{Coherent Loss: A Generic Framework for Stable Video Segmentation}
\author {

         Mingyang Qian\textsuperscript{\rm 1},
         Yi Fu\textsuperscript{\rm 2},
         Xiao Tan\textsuperscript{\rm 2},
         Yingying Li\textsuperscript{\rm 2},
         Jinqing Qi\textsuperscript{\rm 1},
         Huchuan Lu\textsuperscript{\rm 1}, \\
         Shilei Wen\textsuperscript{\rm 2},
         Errui Ding\textsuperscript{\rm 2} \\
}
\affiliations {
    \textsuperscript{\rm 1} Dalian University of Technology,
    \textsuperscript{\rm 2} Department of Computer Vision Technology (VIS), Baidu Inc.
     mingyangqian25@gmail.com,
    fuyi02, tanxiao01, liyingying05, wenshilei, dingerrui@baidu.com, \\
    jinqing, lhchuan@dlut.edu.cn
}

\begin{document}

\maketitle

\begin{abstract}
Video segmentation approaches are of great importance for numerous vision tasks especially in video manipulation for entertainment.
Due to the challenges associated with acquiring high-quality per-frame segmentation annotations and large video datasets with different
environments at scale, learning approaches shows overall higher accuracy on test dataset but lack strict temporal constraints to self-correct jittering artifacts in most practical applications.
We investigate how this jittering artifact degrades the visual quality of video segmentation results and proposed a metric of temporal stability to numerically evaluate it.
In particular, we propose a Coherent Loss with a generic framework to enhance the performance of a neural network against jittering artifacts, which combines with high accuracy and high consistency.
Equipped with our method, existing video object/semantic segmentation approaches achieve a significant improvement in term of more satisfactory visual quality on video human dataset, which we provide for further research in this field, and also on DAVIS and Cityscape.
\end{abstract}

\section{Introduction}

\begin{figure*}[tbp]
\begin{center}
\begin{tabular}{c@{}c@{}c@{}c@{}}\normalsize
\includegraphics[width=0.24\linewidth, height=0.12\linewidth]
{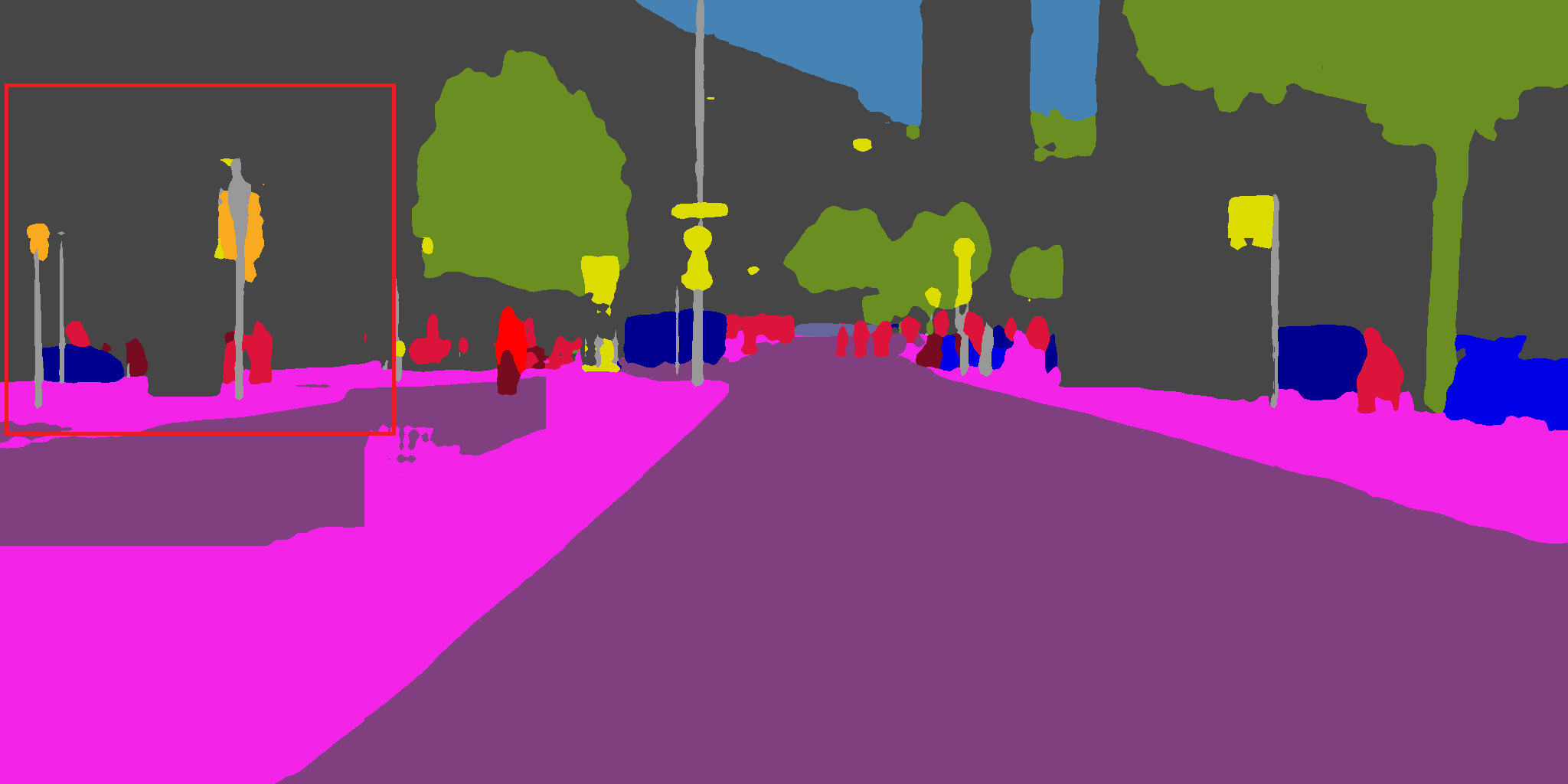} \ &
\includegraphics[width=0.24\linewidth, height=0.12\linewidth]
{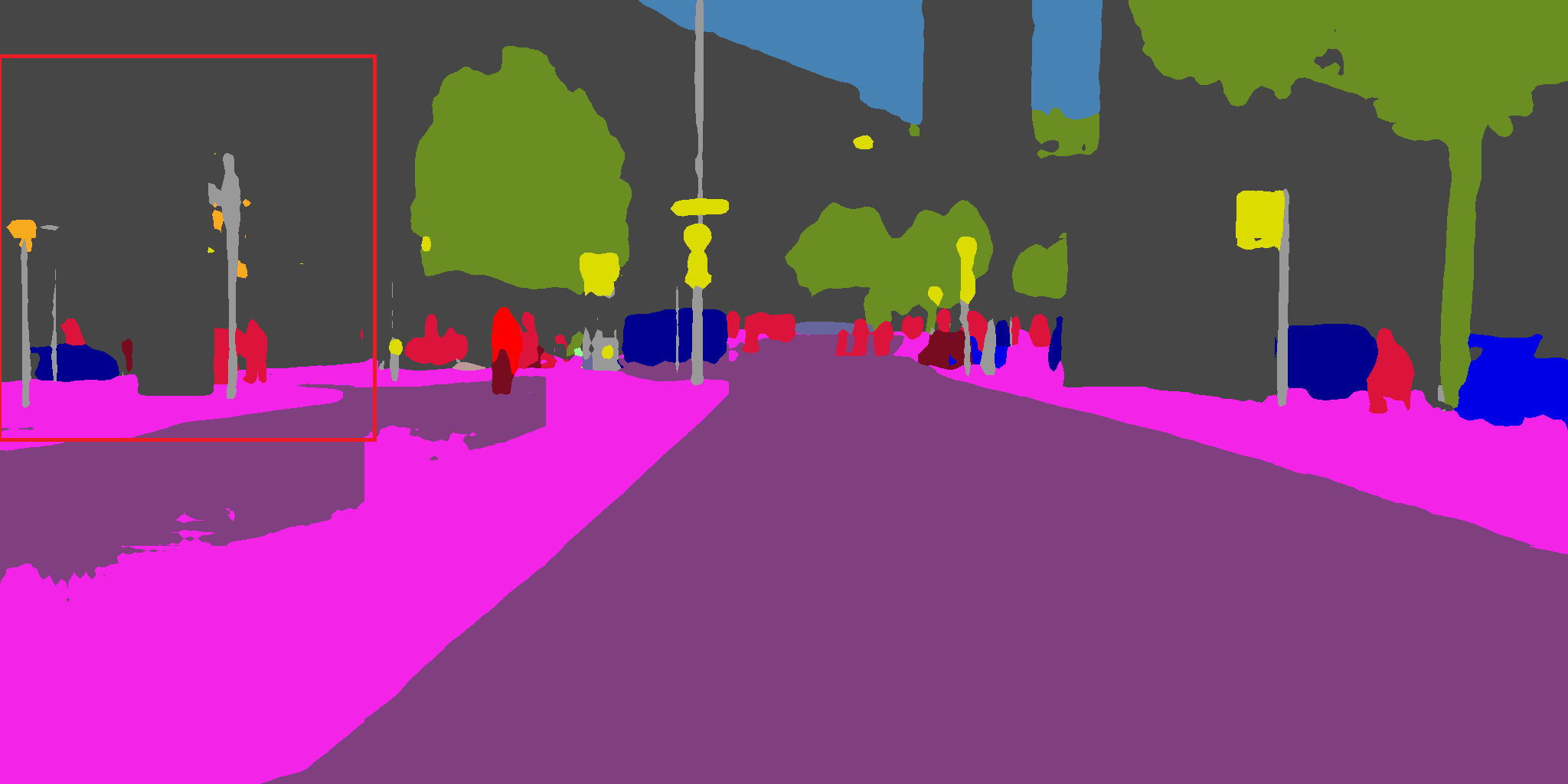} \ &
\includegraphics[width=0.24\linewidth, height=0.12\linewidth]
{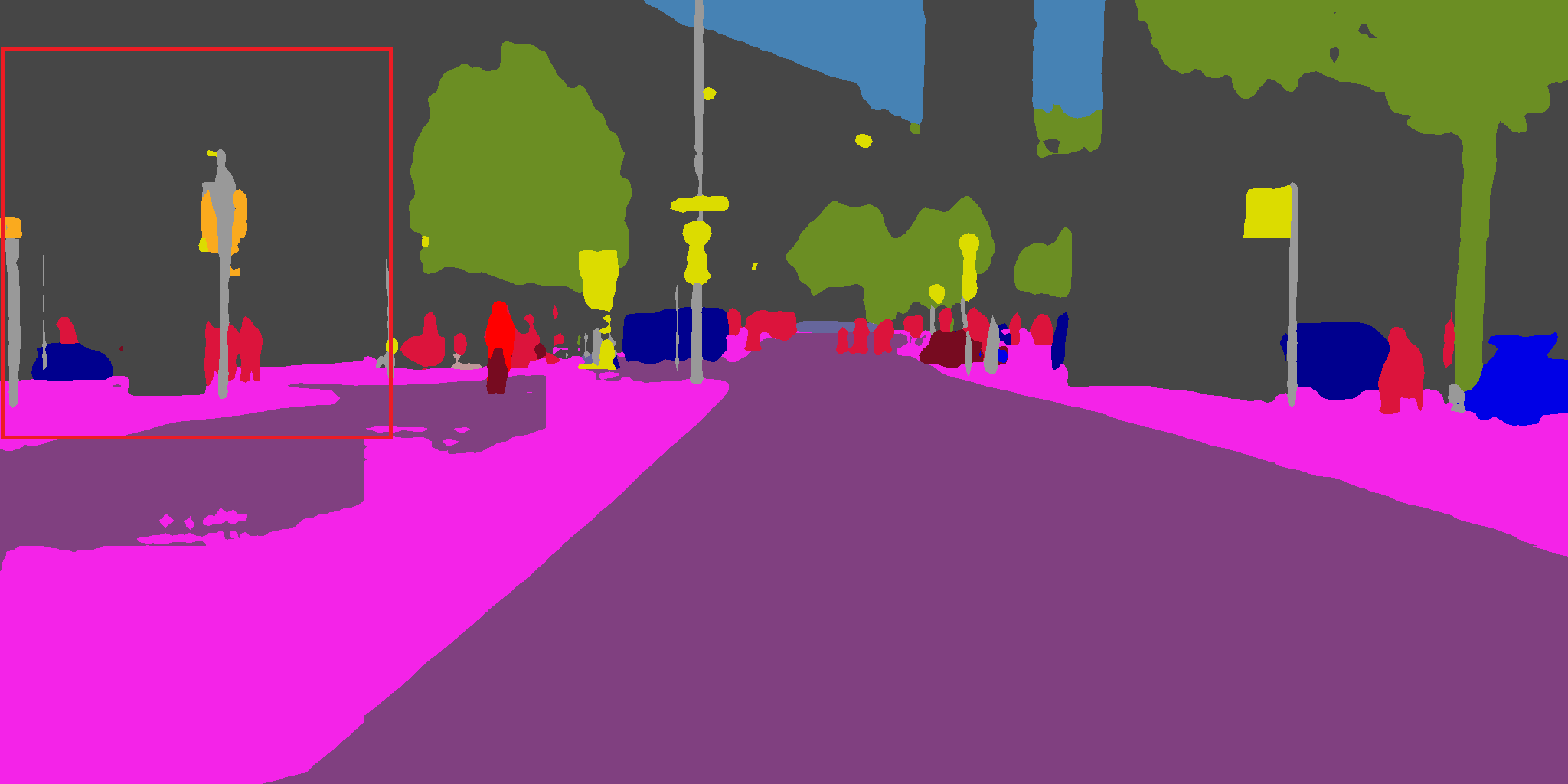} \ &
\includegraphics[width=0.24\linewidth, height=0.12\linewidth]
{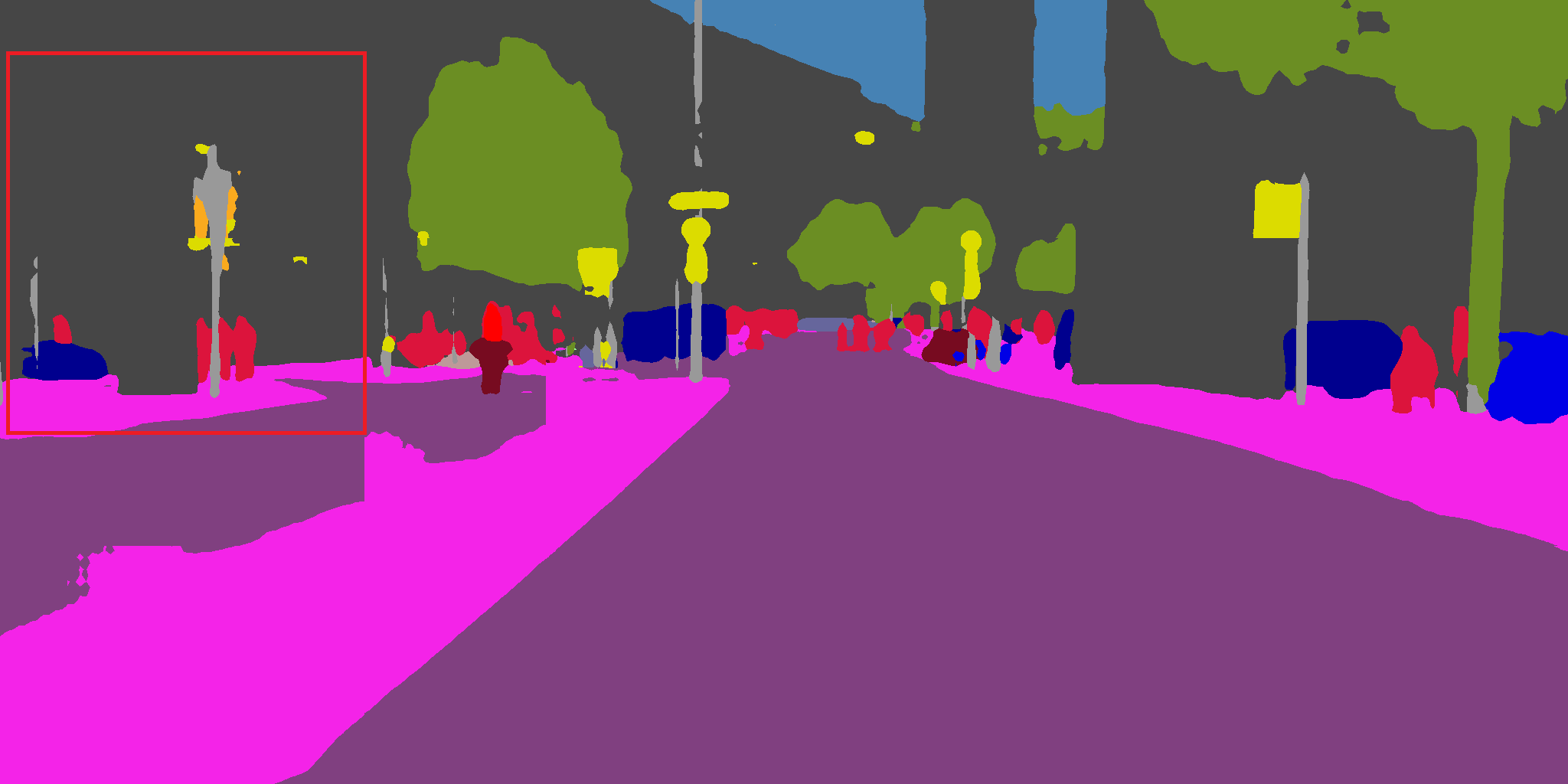} \\
\includegraphics[width=0.24\linewidth, height=0.12\linewidth]
{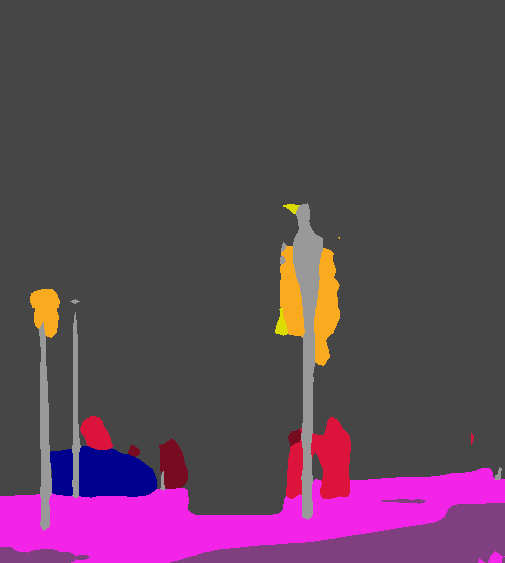} \ &
\includegraphics[width=0.24\linewidth, height=0.12\linewidth]
{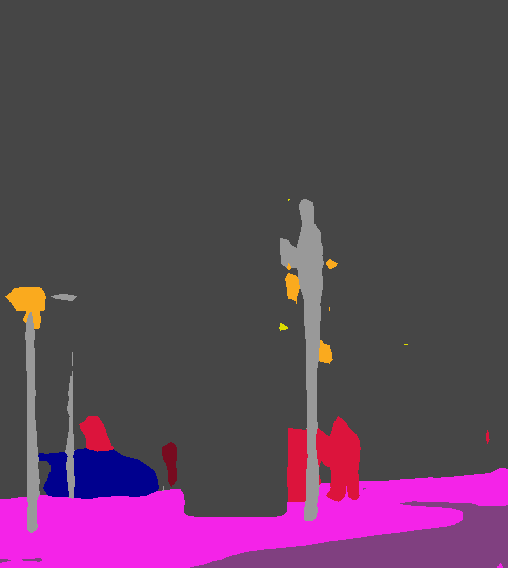} \ &
\includegraphics[width=0.24\linewidth, height=0.12\linewidth]
{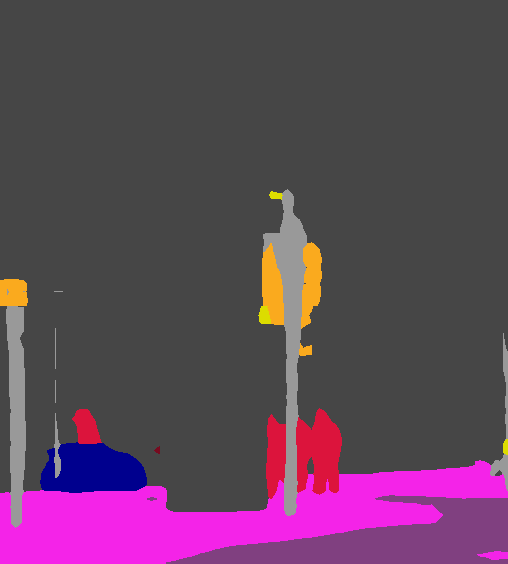} \ &
\includegraphics[width=0.24\linewidth, height=0.12\linewidth]
{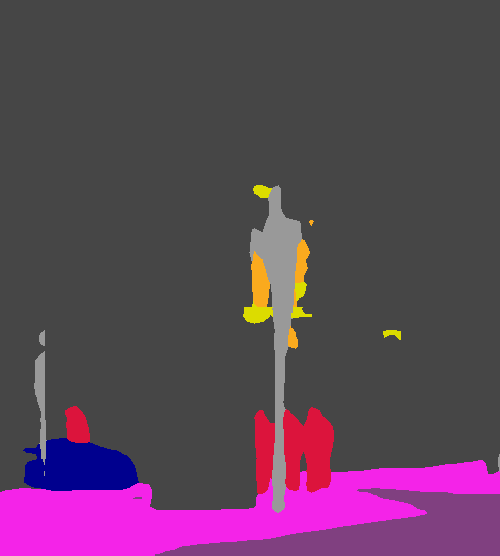} \\
frame t & frame t+1 & frame t+2 & frame t+3 \\
\end{tabular}
\end{center}
\caption{Segmentation results from the official PSPNet~\cite{song2018pyramid}. We give a close-up view for the details in the second row. Note that the inconsistencies appear on the more discriminative results such as the billboard. }
\label{Fig_into}
\end{figure*}

As a dense per pixel-level prediction task, video segmentation aims to label all pixels in a video sequence for semantic segmentation or differentiating individual instance and assigning consistent object IDs to each instance over the sequence for instance segmentation, which remains a foundational building block in many autonomous driving and interactive entertainment products.
For example, in many interactive applications, i.e., background transition, clothes changing and auto clipping, a fast and accurate semantic and/or instance segmentation method is required for targeting the object or human of interest.
Unfortunately, most off-the-shelf methods failed to provide satisfactory video segmentation results, i.e, the results may vary a lot even the scenario undergoes only a tiny change as shown in Figure~\ref{Fig_into}.
This phenomenon could lead to bad alignment and temporal jittering of object boundaries in videos, which significantly undermines visual quality and hence degrades the interactive experience.
In addition, per-pixel accuracy in video segmentation methods like mIoU is used to measure the static segmentation performance of the overall test samples,
but jittering artifact of scattered frames will be propagated to the whole video and generate perceptually jarring results, which cannot reflect real sensory experience through mIoU.

Recent segmentation methods succeed to detect a better object identity with more clear edges, but fail to alleviate the jittering artifact and misaligned distortions along object boundaries when processing images sequentially.
Besides, although temporal coherency has been valued in many methods~\cite{voigtlaender2019feelvos},~\cite{hu2020temporally}, they still face the following unavoidable risks.
Camera motion inevitably leads to motion blurring and confuses the prediction of algorithms, which causes the jittering artifacts.
Even worse is "ground truth" boundaries labeled by annotators are in fact not that accurate enough~\cite{hoebel2019give}, it is known that manual annotations often involve inherently uncertain boundary areas of single image, not to mention annotating perfect alignment across video sequences.
As a result, jittering is observed when we apply a predictor using per-frame annotation independently, and the segmented mask can not adhere well to a complete define object of anatomically in the sequence, even if the masks are annotations.
Besides, metric like mIou that comparing the accuracy of results against the per-frame annotations cannot reflect the temporal coherency when the predictor’s bias is small but variance is large especially along the boundary.
Therefore, we hope to evaluate the jittering artifact as an important indicator, and fundamentally alleviate the jitter of actual results, which is more sensitive for video manipulation applications.

Note that perfect coherency can be achieved trivially at the expense of accuracy: all pixels in the frames are assigned the same label, however, a combination of high accuracy and high coherency is not easy to achieve.
Instead of purely relying on mIoU for evaluation, we first emphasis the temporal coherency and introduce a simple metric: stability rate, to measure the overall prediction consistency in a sequence.
We introduce a Coherent Loss with a generic framework performing as a neural network training strategy smartly exploiting unlabeled videos for enhancing the temporal coherency.
Owing it is orthogonal to the network architecture, it can hence be used to cooperate with state-of-the-art segmentation methods to alleviate the jittering artifacts without adding extra prediction overhead.

Unfortunately, most of the current datasets for segmentation, such as DAVIS-16~\cite{perazzi2016benchmark} and Cityscape~\cite{cordts2016cityscapes}, are small and have a few labeled videos.
Using such a dataset for video segmentation may result in poor performance and be not suitable for evaluating stability.
Besides, the human is a necessary category of segmentation and is easily accessed due to the strong demand in the practical products, thus we collect a large dataset for video human segmentation, which contains about 100,000 labeled images and 130 unlabeled videos for future research. As a result, our contributions are as following:
\begin{itemize}
\item We expose a new dimension of evaluating video segmentation methods in term of temporal coherency which plays an important role in video segmentation tasks, and introduce a new numeric metric, stability rate, to explicitly measure the coherency of video segmentation results.
\item We introduce a Coherent Loss with a generic framework, which performs as a training strategy smartly exploiting unlabeled videos to enhance temporal coherency in video segmentation. Owing it is orthogonal to the network architecture, it can hence be used to cooperate with many state-of-the-art segmentation methods based on neural network without adding extra prediction overhead.
\item We collect a large video human segmentation dataset for research, and we show the improvement of visual stability on this dataset and also achieve promotion on the public DAVIS and Cityscape.
\end{itemize}

\section{Related Work}
\subsection{Segmentation Methods}
Many video object segmentation methods adjust the neural network for better adaption to the propagation of temporal information.
Huang \emph{et al.}~\cite{huang2020fast} propose a new temporal aggregation network and a novel dynamic time-evolving template matching mechanism to achieve significantly improved performance.
David \emph{et al.}~\cite{nilsson2018semantic} propose a Spatio-Temporal Transformer GRU module that can temporally propagate labeling information by FlowNet~\cite{ilg2017flownet}, adaptively gated based on its locally estimated uncertainty.
Wang \emph{et al.}~\cite{wang2019ranet} fuse the features and masks of the previous frame with the current frame and supervise it with current annotation to enhance temporal consistency indirectly.
Though they achieve high performance of mIoU, sequentially running these approaches on each frame of a video usually leads to jittering and unaligned distortions along object boundaries.
Our method is quite different from these work as we strictly constraint the coherency between the tracked pixels by proposed loss function rather than relying on defective labels for per-frame supervision.
In addition, the problem of video jitter as mentioned above cannot be avoided by per-frame supervision (because manual annotation of non-jitter is difficult to obtain), we use unlabeled videos for unsupervised learning to explore temporal stability of the sequential results directly by minimizing pixel-level unmatched errors between these results.

\begin{figure*}[htbp]
\centering
\includegraphics[scale=0.45]{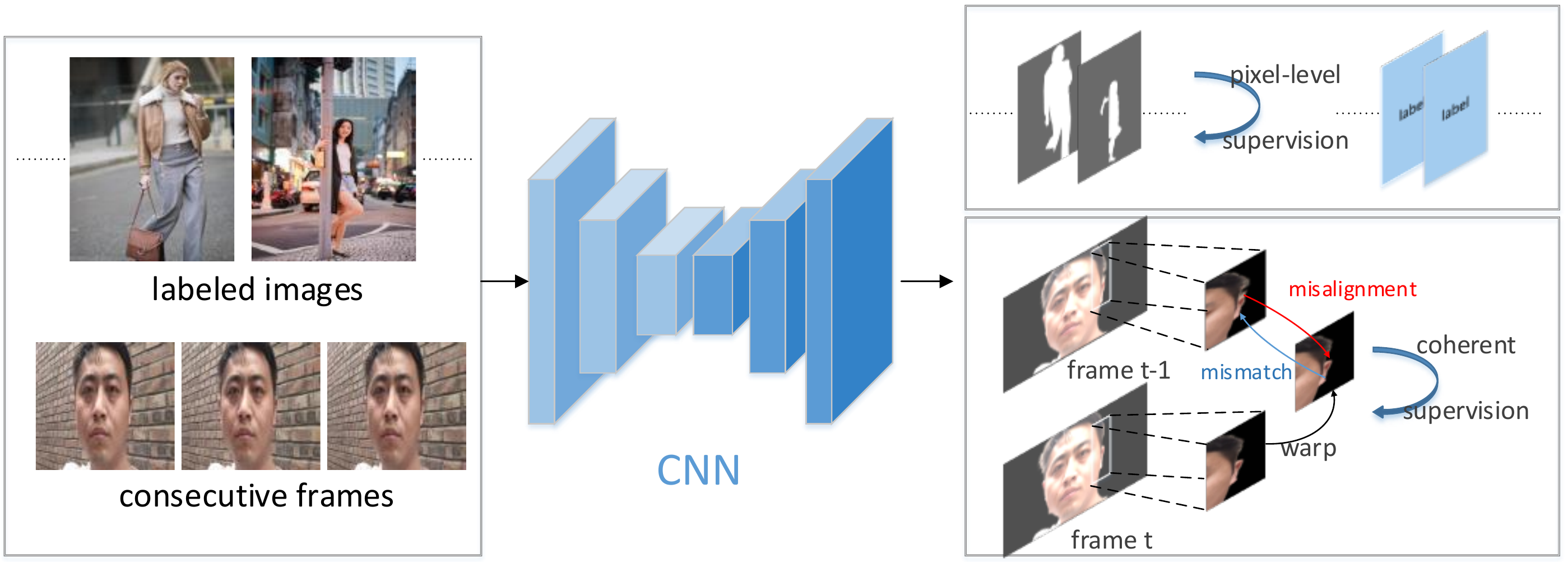}
\caption{Framework overview. The per-pixel supervision utilizes labeled samples for better segmentation while the proposed coherent supervision utilizes unlabeled videos for self-correcting unstable details.}
\label{fig:framework}
\end{figure*}

\subsection{Methods about Coherency}
It is difficult to annotate consistently along the object boundary uncertain areas across frames, thus some methods suggest completely different ideas.
Kundu \emph{et al.}~\cite{kundu2016feature} employ a regularization to optimize the mapping of pixels to a Euclidean feature space so as to minimize distances between corresponding points.
It significantly enhances the temporal coherency after post-processing but is very time-consuming.
In~\cite{dong2018supervision}, supervision-by-registration is proposed to improve jitter of key points in facial landmark detectors on both images and video, which has a great reference to us.
Besides, the metric mIoU, which evaluates the quality of segmentation against the annotation, are also doubted if it can measure the coherency of video segmentation.
Hoebel \emph{et al.}~\cite{hoebel2019give} assess some measures of uncertainty for segmentation in medical imaging and the correlation with segmentation quality, while Hendrycks~\cite{hendrycks2019benchmarking} discusses the robustness of the model under different noise disturbances and provides the metrics on synthetic data sets.
Although the metric consistency in~\cite{kundu2016feature} and~\cite{nilsson2018semantic} is measured if all pixels along the track are assigned the same label, it contains greater error when the positive or negative regions are much larger.
We propose a new metric to explicitly measure the coherency of video segmentation, which is based on the consistent predictions of the corresponding pixels and can selectively reflect the coherency of global or local segmentation in each video, and we also present a novel Coherent Loss to enhance the temporal coherency. 

\section{Method}
\subsection{Framework}
Widely used segmentation loss functions involve a Cross Entropy for per-pixel prediction, which measures some similarity between each prediction and the corresponding ground truth, and may not good at maintaining temporal coherency, we hence introduce our Coherent Loss for alleviating the visual jittering artifact for video segmentation, and the overview of our framework is shown in Figure~\ref{fig:framework}.
We introduce our Coherent Loss for alleviating the visual jittering artifact for video segmentation, and the overview of our framework is shown in Figure~\ref{fig:framework}.
Given a trained network for providing satisfactory video segmentation results, we propose an online matching strategy named coherent supervision to enhance the coherency on the basis of per-pixel supervision.
It is ideally to train with fully labeled video sequences, however the issues are: (1) it is difficult to annotate perfect alignment across video frames and (2) annotating video frames needs huge cost especially for higher accuracy.
Therefore, the proposed Coherent Loss exploits unlabeled videos to learn to predict consistently in corresponding regions between successive frames.
Owing it is orthogonal to the network, this novel paradigm allows us to cooperate with many successful methods for better performance.
\subsection{Per-Pixel Loss}
Many segmentation networks take an image $I$ as input and output a predicted softmax map $M = S(I)$.
Typically, they apply a Cross-Entropy loss on the predicted map $M$ with the ground truth $G$, i.e., $L_{seg} = - \sum_{i=1}^{N} l_{i} log p_{i}$, where $p_{i}$ is the predicted $i\-th$ categorical probability in $M$ while $l_{i}$ is its ground truth labels.
Note that when training with different kinds of networks, we can still use the original loss functions and settings described in their methods.
\subsection{Coherent Loss}
\begin{figure*}[tbp]
\centering
\includegraphics[scale=0.45]{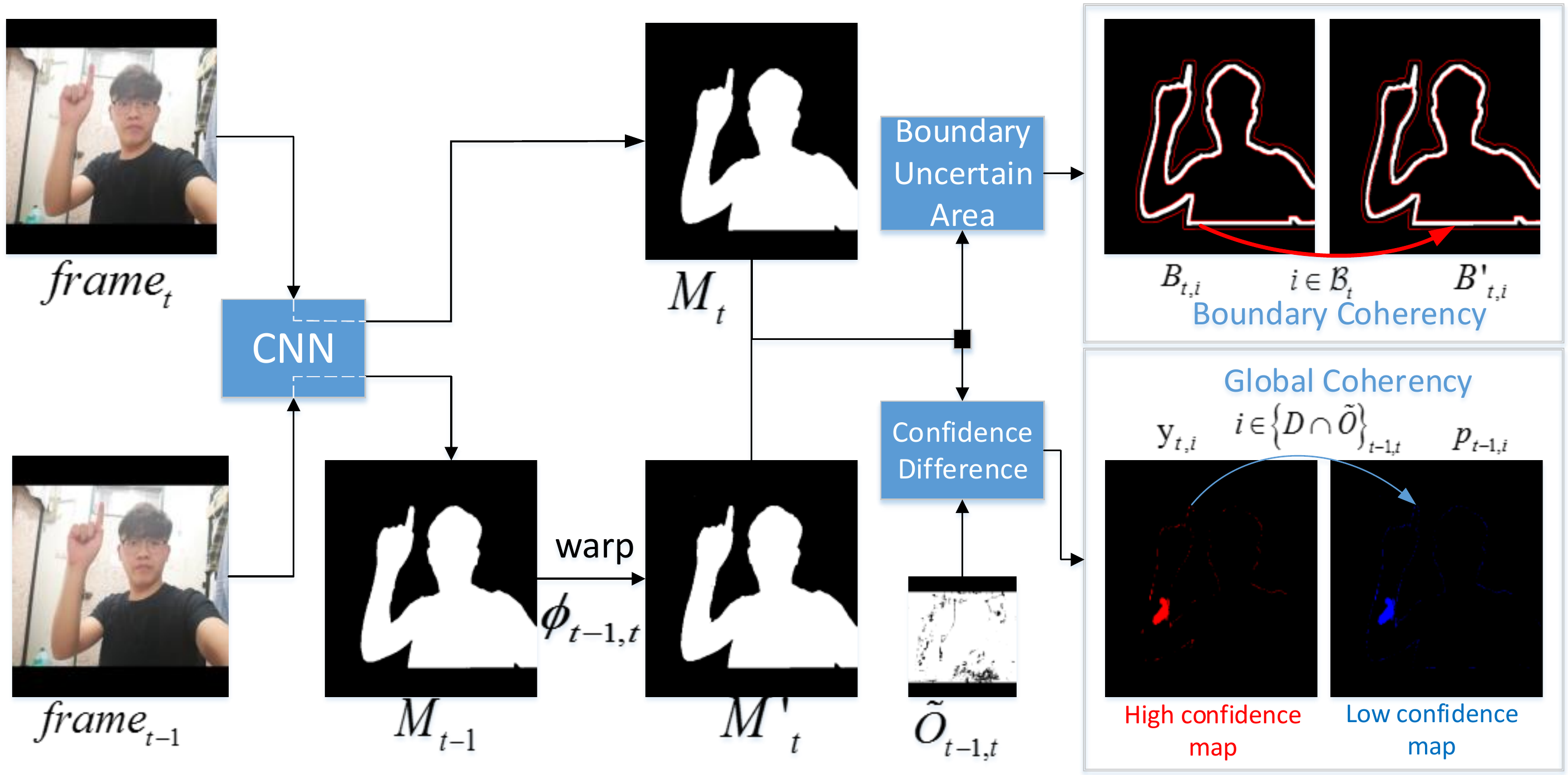}
\caption{The overview of Coherent loss.}
\label{fig:gc}
\end{figure*}

Since temporal coherency requires the corresponding pixels in successive frames having a same label, optical flow is encouraged to find those corresponding pixels in each image pair.
The Coherent Loss is proposed to perform online matching on the unlabeled videos, which directly uses loss function across these corresponding points to calculate mismatched prediction errors and back-propagate gradient for better alignment during the training.
Specifically, Coherent Loss(shown in Figure~\ref{fig:gc}) is required to address two main issues for better temporal coherent results.
The first is misalignment around boundary region caused by training with inconsistent annotations~\cite{hoebel2019give}, thus Boundary Coherency is designed to improve the temporal alignment among the sensitive boundary areas between adjacent frames.
The second is sudden mis-segmentation due to the undetectable appearance difference between different targets, changes in lightening or deformation caused by camera jitter, thus Global Coherency is designed to minimize the mis-segmentation in the whole sequences.

\textbf{Finding Corresponding Pixels.}
Motivated by~\cite{dong2018supervision} who designs a differentiable Lucas-Kanade operation to track facial landmarks, we design an iterative algorithm to calculate the optical flow of the whole image.
We use $\{O\}$ as a set of locations between two frames to indicate all tracked points and $\{\tilde{O}\}$ to indicate those tracked points passed by dual matching of forward-backward check~\cite{kalal2010forward} for better accuracy.
For simplification, we use $\phi_{t-1,t}$ to formulate a mapping of labels in $\{O\}$ or $\{\tilde{O}\}$ along the optical flow, i.e., $\phi( [x,y]^{T} ) = [x+u, y+v]^{T} $, where $(u,v)$ are the offsets.

\textbf{Boundary Coherency.}
As mentioned above, per-frame supervision with imprecise annotations along the boundary uncertain areas leads to incoherency of segmentation boundaries. 
We present Boundary Coherency for the case in which we treat all the corresponding pixels along the boundary areas in the $\{O\}$ to be robust and encouraged to have the same label, enabling the coherency of segmented boundary areas.

We first extract the single line of boundary mask $b_{t}$ calculated by the $M_{t}$ output from the network.
The $b_{t}$ is considered as the center of the boundary uncertain area, we extend the boundary line $b_{t}$ to the area with a certain width, covering the potential areas of uncertainty.
We set the width to several pixels in accordance with general conditions, the boundary uncertain area mask is defined as ${B}_{t} = extend( b_{t}, \theta)$, which is a binary map.
Then, all the positions in the uncertain areas, donates as $\{\mathcal{B}_{t}\} = \{ i | {B}_{t,i} = 1 \}$, are regarded as the hypothetical ground truth to supervise the corresponding predictions in the $M^{'}_{t}$, warped from the $M_{t-1}$ by $M^{'}_{t}(x^{'},y^{'}) = M_{t-1}(\phi_{t-1,t}(x,y))$.
We use the standard Lovasz-Softmax loss($LS$) for the optimization of the intersection-over-union just in $\{\mathcal{B}_{t}\}$.
Similarly, the backward part also performs the same processing, thus the Boundary Coherency loss ($L_{bc}$) is defined as following:
\begin{equation}
\label{Boundary_Consistency}
    \begin{split}
         L_{bc} =
         &\frac{1}{2}{\sum}_{i\in\mathcal{B}_{t}} LS \left(M^{'}_{t,i}, \bar{l}_{t,i}\right) + \\
         &\frac{1}{2}{\sum}_{i\in \mathcal{B}_{t-1}} LS \left(M^{'}_{t-1,i}, \bar{l}_{t-1,i}\right).
    \end{split}
\end{equation}
$\bar{l}_{t-1,i}$ and $\bar{l}_{t,i}$ are the hypothetical ground truth labels from the $M_{t-1}$ and $M_{t}$ at $i\-th$ position.

\textbf{Global Coherency.}
As mentioned above, mis-segmentation due to sudden changes in external environment brings temporal jittering in the global scope.
We handle it by introducing the learning of temporal information to keep the prediction consistency among $M$ from a global perspective.
Therefore, the motivation of Global Coherency is that the predicted $M_{t}$ will perform as a reference for the $M_{t-1}$ to learn to get closer to this result.

As shown in the Figure~\ref{fig:gc}, from $frame_{t-1}$ to $frame_{t}$, we use the prepared optical flow to warp the $M_{t-1}$ into the next time, by computing $M^{'}_{t}(x^{'},y^{'}) = M_{t-1}(\phi_{t-1,t}(x,y))$.
As Global Coherency $L_{gc}$ serves as an unsupervised manner, the model training may suffer from the misalignment by directly using labels from the $M_{t}$ for supervision.
But, those labels with higher confidence scores should be trusted more than those with lower confidence scores.
Thus, we adopt a compromise plan that we find out all the points with higher top-1 scores in the $M_{t}$ than those aligned in the $M^{'}_{t}$ through an appropriate threshold $\gamma$, which denotes as $\{D\}_{t-1,t} = \{ i | | M_{t,i} - M^{'}_{t,i} | \textgreater \gamma  \}$.

To be more accurate, we use $\{D\cap \tilde{O}\}_{t-1,t}$ to denote these tracked points with higher confidence scores and Soft Cross-Entropy~\cite{de2005tutorial} loss to reduce the mismatched errors in these tracked pixels.
Similarly, we also perform the Global Coherency from $frame_{t}$ to $frame_{t-1}$, thus the loss is defined as following:
\begin{equation}
\label{Global_Consistency}
    \begin{split}
        L_{gc} =
        &-\frac{1}{2} {\sum}_{i\in\left\{D\cap \tilde{O}\right\}_{t-1,t}} y_{t,i} * log {p^{'}_{t,i}} \\
        &-\frac{1}{2} {\sum}_{i\in \left\{ D\cap \tilde{O}\right\}_{t,t-1}} y_{t-1,i} * log {p^{'}_{t-1,i}}.
    \end{split}
\end{equation}
$y_{t-1,i}$ and $y_{t,i}$ denote the target softmax scores at $i\-th$ position in $M_{t-1}$ and $M_{t}$, and $p^{'}_{t-1,i}$ and $p^{'}_{t,i}$ denote the predicted scores in the warped $M^{'}_{t-1}$ and $M^{'}_{t}$.

\begin{table*}[tbp]
\begin{center}
\caption{User-study for real human videos. For simplification, Method1 and Method2 are the base model in the ~\cite{li2019self} with ResNet18 and ResNet50 as backbone; Method3, Method4, Method5 are fine-tuned on the trained model of Method2, with Coherent Loss of different parameters using unlabeled videos.}
\label{table:user}
\begin{tabular}{cccccccc}
\hline
  & stable videos & score & mean score & ${STB}_{global}$     & ${STB}_{local}$ & mIoU(syn) & mIoU(test)\\
\hline
Method1 &  7 / 30  &  68  & 2.27       & 99.17                & 95.46           & 91.20     & 95.14\\
Method2 &  8 / 30  &  68  & 2.27       & 99.18                & 95.54           & 93.22     & 96.20\\
Method3 &  17/ 30  &  76  & 2.53       & 99.22                & 95.94           & 93.80     & 95.37\\
Method4 &  27/ 30  &  91  & 3.03       & 99.28                & 96.54           & 94.28     & 94.28\\
Method5 &  26/ 30  &  95  & 3.17       & 99.29                & 96.79           & 94.31     & 95.91\\
\hline
\end{tabular}
\end{center}
\end{table*}

\subsection{Total Loss}
The complete loss function contains two important components: per-pixel segmentation loss $L_{seg}$ for spatial supervision, and Boundary Coherency $L_{bc}$ and Global Coherency $L_{gc}$ for coherent supervision:
\begin{equation}
    \label{all_Loss}
        L_{all} = L_{seg} + \alpha L_{bc} + \beta L_{gc}.
\end{equation}
We use a balanced combination of the segmentation loss and coherent loss controlled by the parameters $\alpha$ and $\beta$.

\subsection{Dataset}
We propose a large video human segmentation dataset (VHS) consisting of labeled images and unlabeled videos.
The labeled images are collected from the public, which contain about 5168 multi-person images from LIP~\cite{gong2017look}, chosen as the completely annotated human instances, 17706 images from ATR~\cite{liang2015human}, 34426 images from AISegment~\cite{AISegment}, about 38781 images from AIC~\cite{wu2019large} and 5200 images on Supervisely Person Dataset~\cite{SPD}.
We relabeled all static images to binary mask, and the unlabeled videos are selfie videos that we have collected from the different people, lasting from 30 seconds to 1 minute.
Details will be described in the Supplementary Material.

\section{Experiments}
We propose a new metric for evaluating the temporal coherency of each method, and apply it to different models.

\subsection{Metric}
A more satisfactory video segmentation means that for a particular pixel (except for points exceeding the image scope) in the video, the model predicts a same and accuracy label across all frames.
We propose a new metric: stability rate (${STB}$) tracking each particular pixel and measuring the coherency of its predictions during the whole sequences.
However, it is too difficult to track the offset of a pixel through annotations, we have to find some finely labeled images to generate video sequences for evaluation using synthesis.

According to~\cite{hendrycks2019benchmarking}, we choose 4 common perturbations: translation, rotation, scaling and occlusion, and 1 corruption: Gaussian noise with fixed parameters, to generate the synthetic test dataset.
Each perturbation type is set to six levels of severity to simulate real-world perturbations.
For each severity, we mark the labeled image as the first image and further randomly perform one of the perturbations to generate the second image, the following frames in the sequence is also a perturbation of the previous with minute Gaussian noise applied all along the line.
We set the length of a sequence of 11 (including the first image) to make sure repeated application of a perturbation does not bring the image far out-of-distribution, so the total synthetic images are $6\times11$ for each image.

Since the next frame is generated from previous frame, the matching pair or optical flow $\{\bar{O}\}_{t-1,t}$ is known for all pixels.
Besides, as the ground truth flow is given, we can calculate the stability in every specific region even the whole image (donated as $\omega$) by using part or all locations in $\{\bar{O}_{\omega}\}_{t-1,t}$, therefore, all un-occluded pixels can be used for evaluation.
For this propose, the ${STB}_{\omega}$ is a rate of pixels in specific region with consistent predictions among all un-occluded pixels.
To be more realistic, we use the generic optical flow $\{O\}_{t-1,t}$ to limit the scope of the evaluation.
In these tracked points in $\{ \bar{O}_{\omega}\cap O\}_{t-1,t}$, the ${STB}_{\omega}$ calculates the proportion of pixels with consistent predictions tracked from $M_{t-1}$ to $M_{t}$ in each two successive frames.
For notation brevity, we assume there is only one video or sequence with N frames.
The equation is defined as:
\begin{equation}
\label{stb}
    \begin{split}
        {STB}_{\omega} =
        &\left( \frac{ 1 }{ N-1 } \sum_{t=2}^{N} \frac{ {\sum}_{i\in\{ \bar{O}_{\omega}\cap O\}_{t-1,t}} (\bar{l}^{'}_{t,i} == \bar{l}_{t,i})} {| \{ \bar{O}_{\omega}\cap O \}_{t-1,t} |} \right) \\
        &* 100\%.
   \end{split}
\vspace{-2mm}
\end{equation}
$\bar{l}^{'}_{t,i}$ or $\bar{l}_{t,i}$ denotes the predicted label from the $M^{'}_{t}$ or $M_{t}$ at $i\-th$ position;
N is the length of each video sequence in the synthetic test dataset.

To make sure the proposed $STB$ is a valid metric for measuring the coherency and it is consistent with the human perception, we carry out the user-study on VHS, which is more common and suitable for observation and evaluation.
We recruited 30 students to rank 5 methods based on results from 30 videos with different ${STB}_{global}$(works on the whole region) and ${STB}_{local}$(works on boundary area with 15-pixel width of ground truth edge).
For each segmented video, we set four evaluation scores: 1 point for inaccurate and unstable segmented video, 2 point for accurate but unstable segmented video, 3 point for inaccurate but stable segmented video, and 4 point for accurate and stable segmented video.
All ratings are subjective evaluations, and we recorded the number of videos considered to be stable, the score of all video and the average score.

We use ${STB}_{global}$, ${STB}_{local}$ and mIoU(syn) on the synthetic test dataset, and mIoU(test) on the test set for reference.
Table~\ref{table:user} shows the results, for the ${STB}_{local}$  of 95.46 on Method1, there are only 7 videos are regarded as stable segmentation, and the average score is 2.27 which means that users think the method is accurate but quite unstable.
With the model ${STB}$ rises, more videos are regarded to be stably segmented.
When the ${STB}_{global}$ rises up to 96.79, mean score achieves the best 3.17 with about 87\% videos are regarded as stable segmentation.
However, one may concern that the accuracy of the method provides a stable foundation, we find even Method2 improves the mIoU(syn) by 2\% compared with Method1, the mean score is still 2.27 that users don't think there is a significant improvement.
Therefore, we believe that the stability metric reveals the intuitive human perception, which verifies the rationality of the metric.

\begin{figure*}[htbp]
\begin{center}
\begin{tabular}{c@{}c@{}c@{}c@{}c@{}c@{}c@{}}\small
\rotatebox{90}{\centering{\small{SCHP}}} \ &
\includegraphics[width=0.15\linewidth, height=0.10\linewidth]
{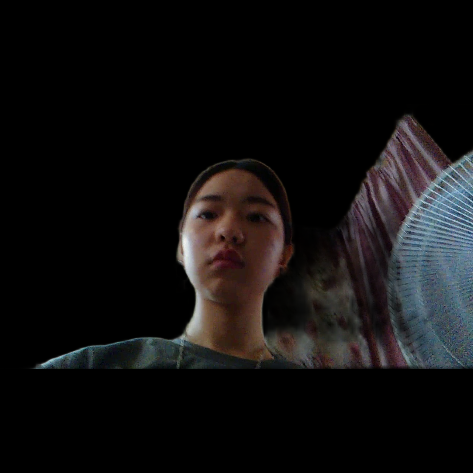} \ &
\includegraphics[width=0.15\linewidth, height=0.10\linewidth]
{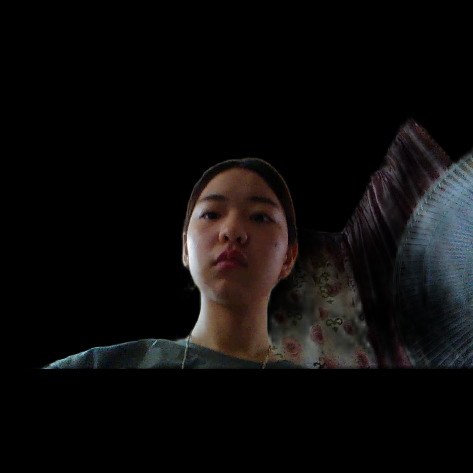} \ &
\includegraphics[width=0.15\linewidth, height=0.10\linewidth]
{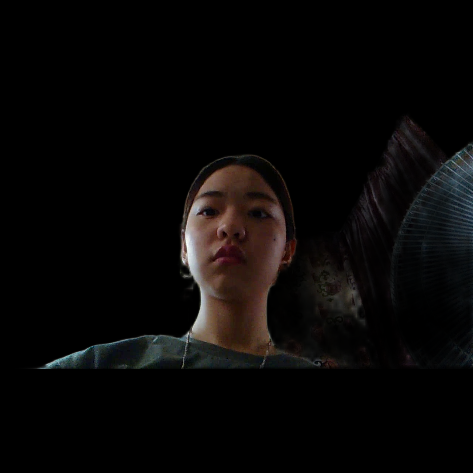} \ &
\includegraphics[width=0.15\linewidth, height=0.10\linewidth]
{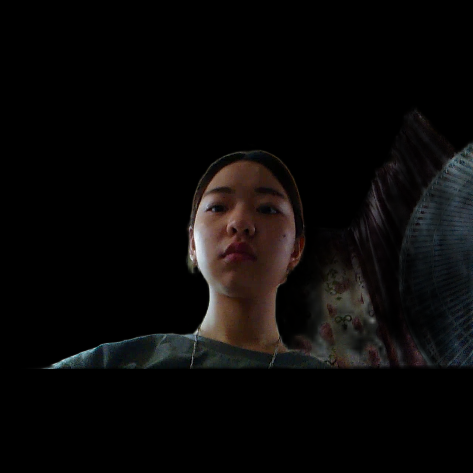} \ &
\includegraphics[width=0.15\linewidth, height=0.10\linewidth]
{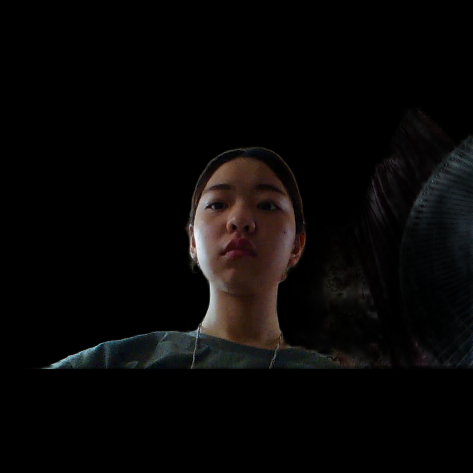} \ &
\includegraphics[width=0.15\linewidth, height=0.10\linewidth]
{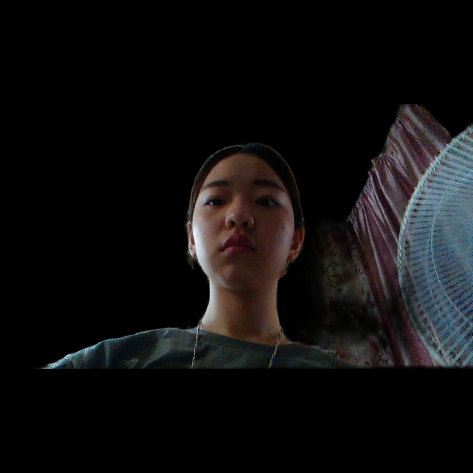} \\
\rotatebox{90}{\centering{\small{SCHP+CL}}} \ &
\includegraphics[width=0.15\linewidth, height=0.10\linewidth]
{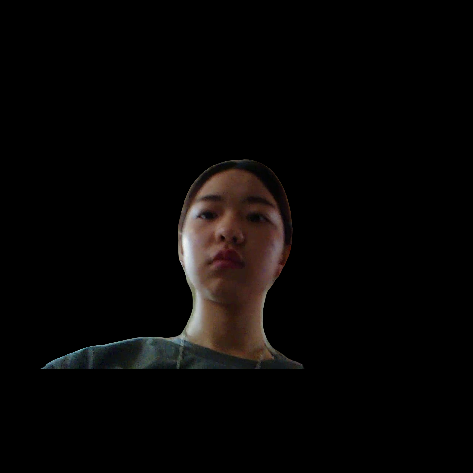} \ &
\includegraphics[width=0.15\linewidth, height=0.10\linewidth]
{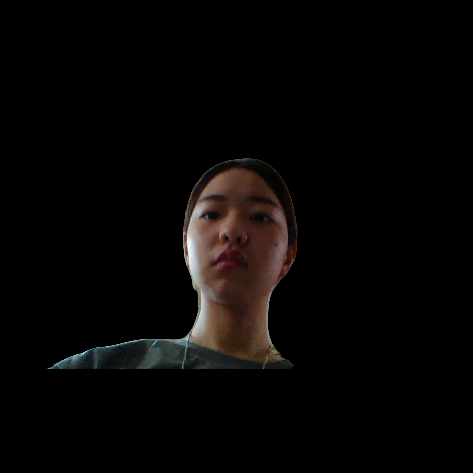} \ &
\includegraphics[width=0.15\linewidth, height=0.10\linewidth]
{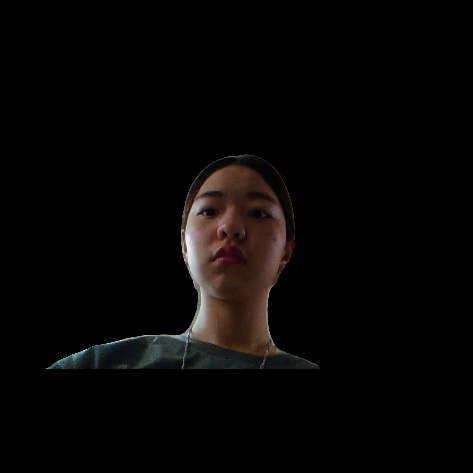} \ &
\includegraphics[width=0.15\linewidth, height=0.10\linewidth]
{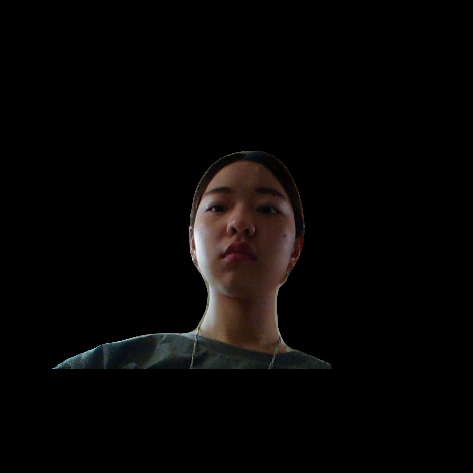} \ &
\includegraphics[width=0.15\linewidth, height=0.10\linewidth]
{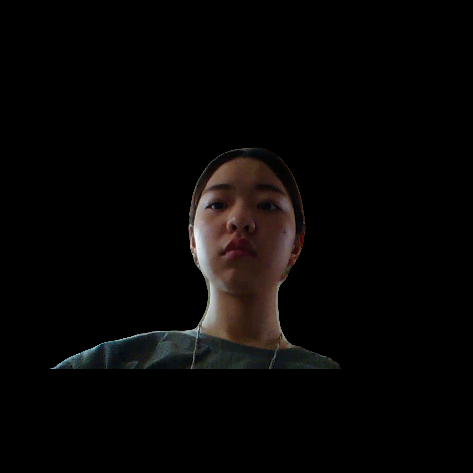} \ &
\includegraphics[width=0.15\linewidth, height=0.10\linewidth]
{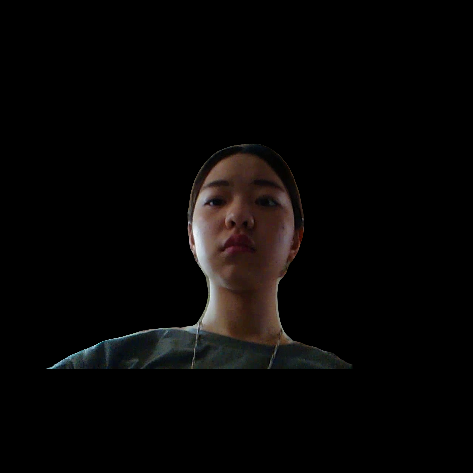} \\
\rotatebox{90}{\centering{\small{SCHP}}}  \ &
\includegraphics[width=0.15\linewidth, height=0.10\linewidth]
{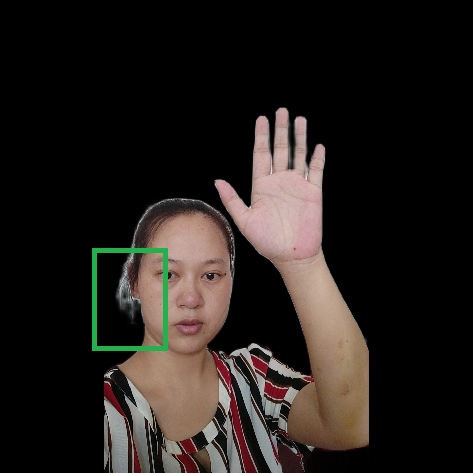} \ &
\includegraphics[width=0.15\linewidth, height=0.10\linewidth]
{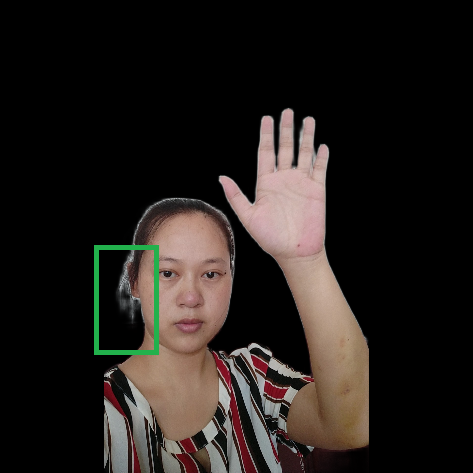} \ &
\includegraphics[width=0.15\linewidth, height=0.10\linewidth]
{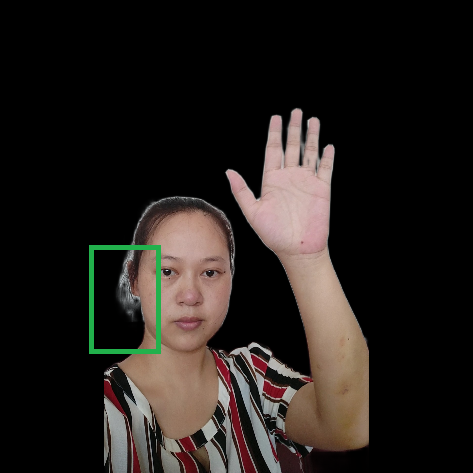} \ &
\includegraphics[width=0.15\linewidth, height=0.10\linewidth]
{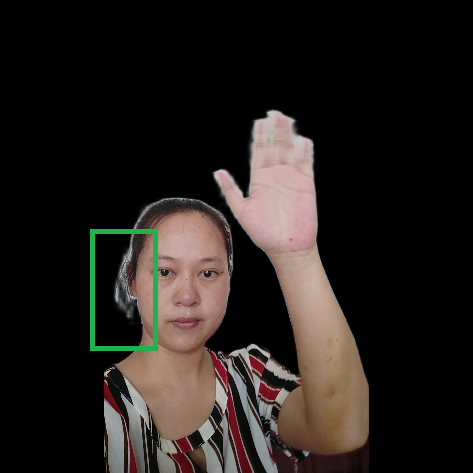} \ &
\includegraphics[width=0.15\linewidth, height=0.10\linewidth]
{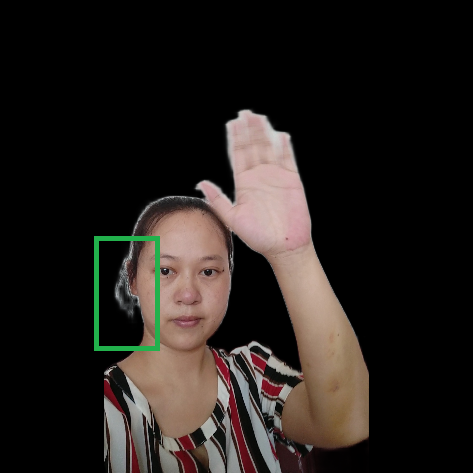} \ &
\includegraphics[width=0.15\linewidth, height=0.10\linewidth]
{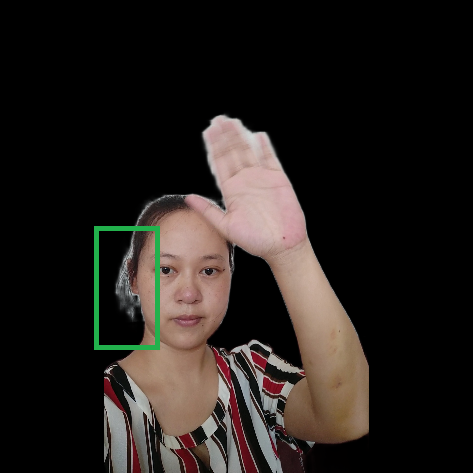}  \\
\rotatebox{90}{\centering{\small{SCHP+CL}}}  \ &
\includegraphics[width=0.15\linewidth, height=0.10\linewidth]
{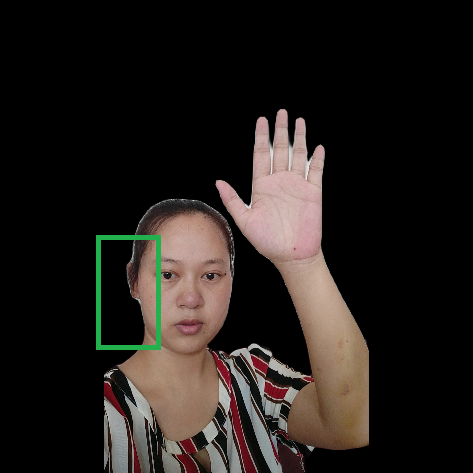} \ &
\includegraphics[width=0.15\linewidth, height=0.10\linewidth]
{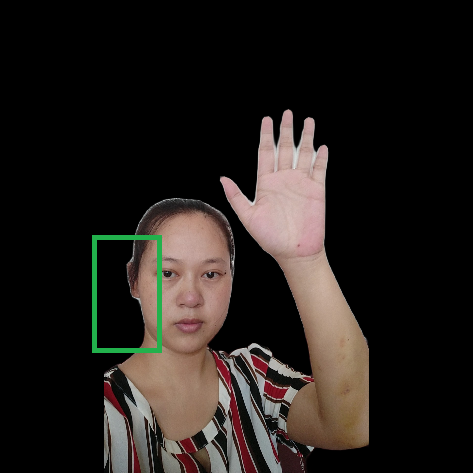} \ &
\includegraphics[width=0.15\linewidth, height=0.10\linewidth]
{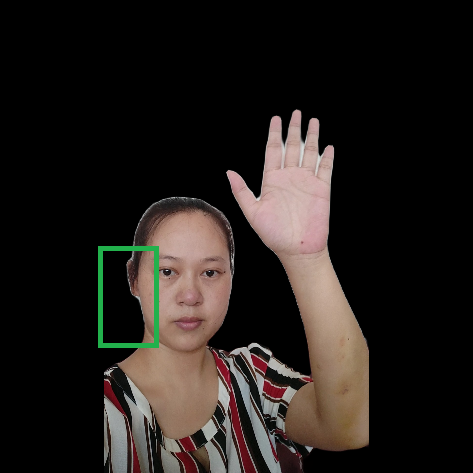} \ &
\includegraphics[width=0.15\linewidth, height=0.10\linewidth]
{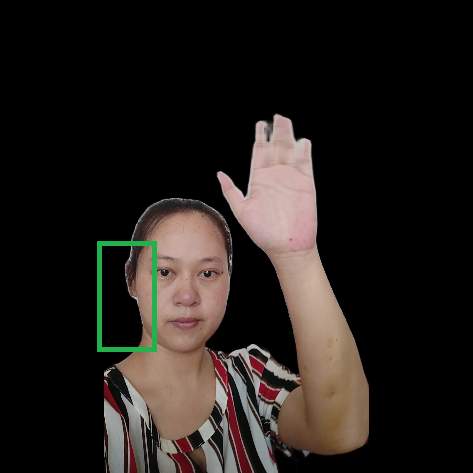} \ &
\includegraphics[width=0.15\linewidth, height=0.10\linewidth]
{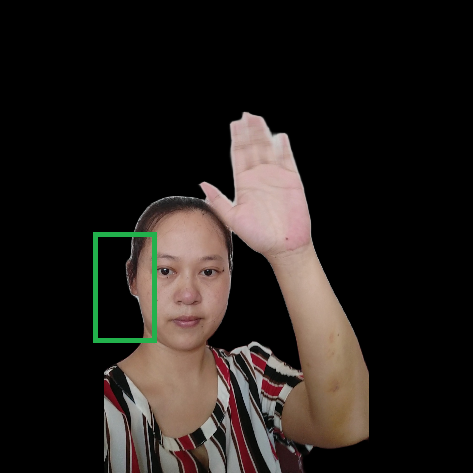} \ &
\includegraphics[width=0.15\linewidth, height=0.10\linewidth]
{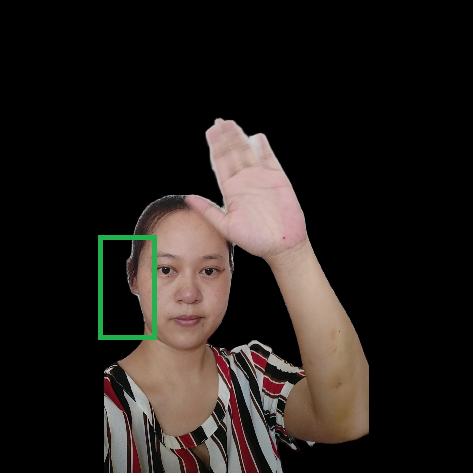} \\
& frame $t$ & frame $t+1$ &frame $t+2$ &frame $t+3$ &frame $t+4$ & frame $t+5$
\end{tabular}
\end{center}
\caption{Qualitative video results for comparison of the method SCHP with/without CL on VHS. We show significant improvement for stable results across adjacent frames.}
\label{Fig_results}
\end{figure*}

\subsection{Video Human Segmentation}
\textbf{Dataset.}
Video human segmentation dataset(VHS) provides 101,281 labeled images for human body and 130 unlabeled videos.
We leverage 91,281 images and 100 videos for training, and 10,000 images and 30 videos for testing.

\textbf{Experiment Settings.}
We exploit the SCHP implementation by~\cite{li2019self} as our baseline and further perform our Coherent Loss on it.
SCHP is designed to enhance the boundary accuracy for improving the accuracy of the whole mask in person-part parsing, which is a state-of-the-art method that ranks 1st in CVPR2019 LIP Challenge.
We adjust this network by using ResNet50 as backbone and replacing the output layer for binary segmentation.

The training procedure is divided into two step.
We first use the ImageNet pre-trained weights to initial the network and train the baseline on labeled images with balanced Cross-Entropy and Lovasz-Softmax loss.
The input labeled images are resized into $473 \times 473$ with data augmentation, and the batch size is 24 for 200 epochs per GPU on Tesla P40.
We use SGD optimizer with an initial learning rate of 7e-3. Momentum and weight decay are set to 0.9 and 5e-4 respectively.
We first make sure the segmentation loss has converged before activating the Coherent Loss.
Second, we use unlabeled videos to fine-tune the network with the proposed loss function.
We sample each video at equal intervals to generate about 30 image pairs of two consecutive frames per video, and all the images are resized into $473 \times 473$.
The training batch size is set to 24 for total 10 epoches, and for each batch we use 12 images from the labeled images for full supervised and 6 random image pairs for coherent supervision.
We use SGD optimizer with an initial learning rate of 1e-4. Momentum and weight decay are set to 0.9 and 5e-4 respectively.
In the equation~\ref{Boundary_Consistency}, the width $\theta$ of the boundary uncertain areas is set to 15 and the $\gamma$ for global consistency is 5e-2 according to our research.
The weight $\alpha$ for boundary coherency loss (bc) in equation~\ref{all_Loss} is set to 1 and the $\beta$ for global coherency loss (gc) is set to 5e-5, enabling a balance between the supervised loss and Coherent Loss.

The metrics of the evaluation are the mIoU on the VHS test set, and ${STB}_{global}$ calculated between global successive masks and ${STB}_{local}$ calculated along the boundary areas (15-pixel width of ground truth edge) on the synthetic test set.
The first frame for the synthesis is randomly collected from the videos for human segmentation on the DAVIS dataset for better annotations.
All the synthetic sequences are 228 with 11 frames per sequence.

\begin{table*}[tbp]
\begin{center}
\caption{Comparison of mIoU on VHS test set, mIoU and $STB$ on synthetic test set.}
\label{table:vhd}
\begin{tabular}{ccccc}
\hline
 Method                         & mIoU(test)  & mIoU(syn) & ${STB}_{global}$   & ${STB}_{local}$ \\
\hline
SCHP(ResNet18)                  & 95.14       & 91.20     & 99.17         &  95.46   \\
SCHP(ResNet18)+CL               & 95.06       & 91.86     & 99.17         &  96.37   \\
SCHP(ResNet50)                  & 96.20       & 93.22     & 99.18         &  95.54   \\
SCHP(ResNet50)+bc($\beta=0$)    & 94.28       & 94.28     & 99.28         &  96.54   \\
SCHP(ResNet50)+gc($\alpha=0$)   & 95.37       & 93.80     & 99.22         &  95.59   \\
SCHP(ResNet50)+CL               & 95.91       & 94.31     & 99.29         &  96.94   \\
\hline
\end{tabular}
\end{center}
\end{table*}

\begin{table*}[tbp]
\begin{center}
\caption{Comparison of mIoU on DAVIS val set, ${STB}$ and mIoU on the synthetic test set.}
\label{table:headings}
\begin{tabular}{cccccc}
\hline
  Method      & mIoU(val) & mIoU(val)  & mIoU(syn) &${STB}_{global}$   &${STB}_{local}$ \\
\hline
        & offline    &  online      & offline &  offline  &offline \\
CRN     &   73.38    &  82.32       &  81.30  & 99.07     & 93.87  \\
CRN+CL  &   75.01    &  83.03       &  82.03  & 99.21     & 94.52  \\
SCHP    &   70.66    &  78.99       &  66.77  & 98.31     & 93.42  \\
SCHP+CL &   75.51    &  81.59       &  69.52  & 98.79     & 94.25  \\
\hline
\end{tabular}
\end{center}
\end{table*}

\begin{figure*}[tbp]
\begin{center}
\begin{tabular}{c@{}c@{}c@{}c@{}c@{}c@{}c@{}}\small
\rotatebox{90}{\centering{\small{CRN}}}  \ &
\includegraphics[width=0.15\linewidth, height=0.10\linewidth]
{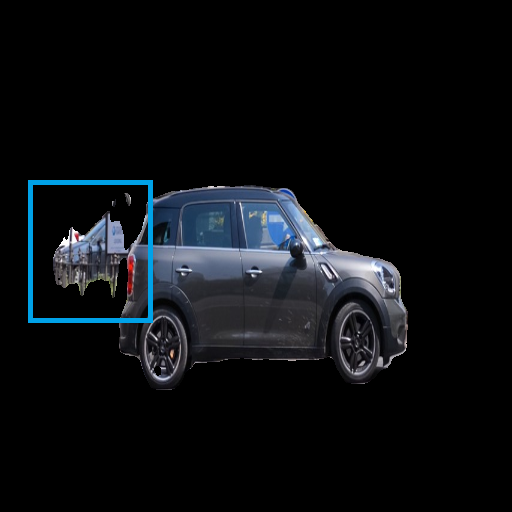} \ &
\includegraphics[width=0.15\linewidth, height=0.10\linewidth]
{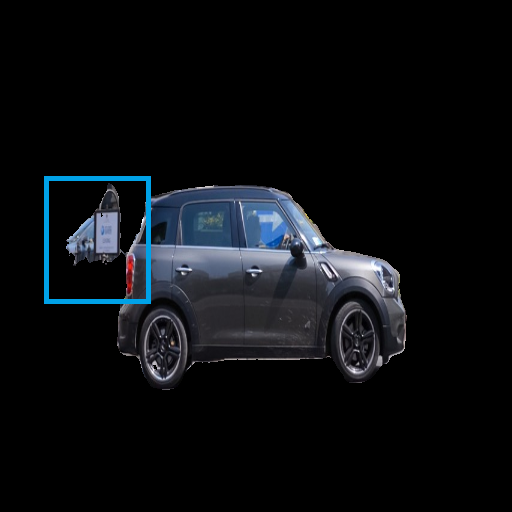} \ &
\includegraphics[width=0.15\linewidth, height=0.10\linewidth]
{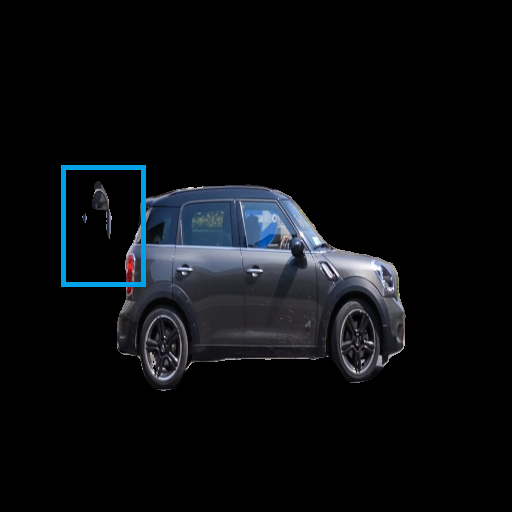} \ &
\includegraphics[width=0.15\linewidth, height=0.10\linewidth]
{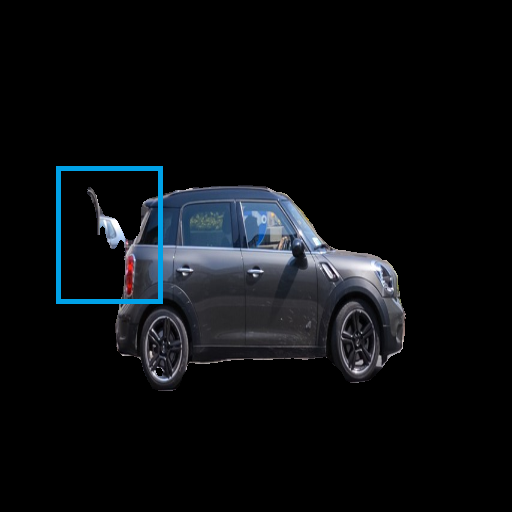} \ &
\includegraphics[width=0.15\linewidth, height=0.10\linewidth]
{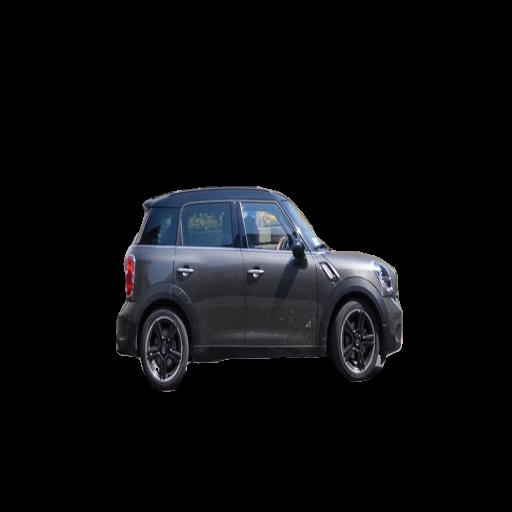} \ &
\includegraphics[width=0.15\linewidth, height=0.10\linewidth]
{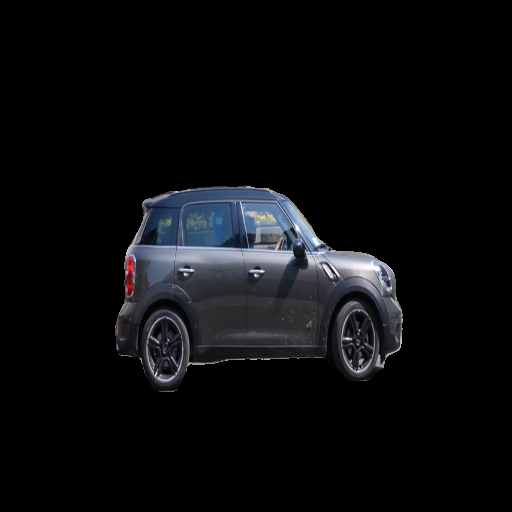} \\
\rotatebox{90}{\centering{\small{CRN+CL}}}  \ &
\includegraphics[width=0.15\linewidth, height=0.10\linewidth]
{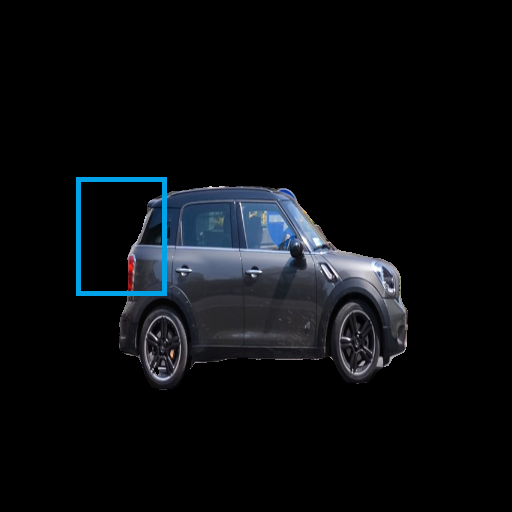} \ &
\includegraphics[width=0.15\linewidth, height=0.10\linewidth]
{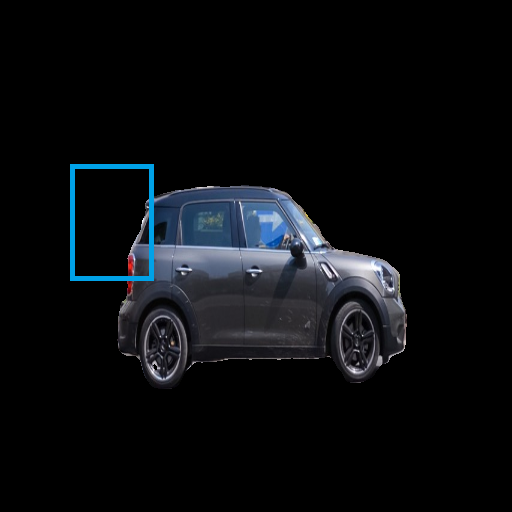} \ &
\includegraphics[width=0.15\linewidth, height=0.10\linewidth]
{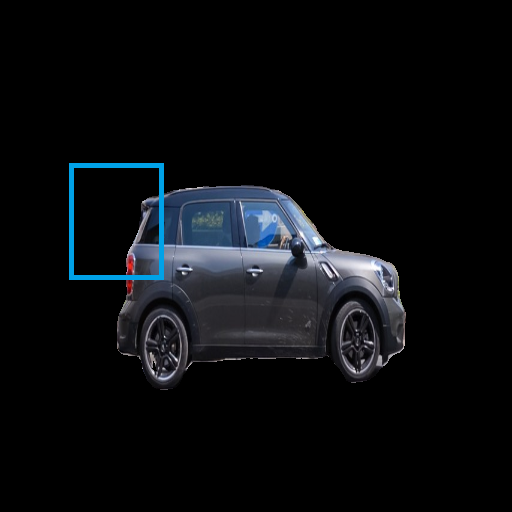} \ &
\includegraphics[width=0.15\linewidth, height=0.10\linewidth]
{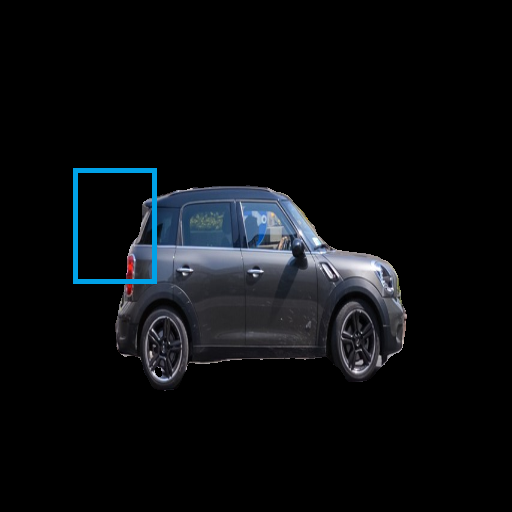} \ &
\includegraphics[width=0.15\linewidth, height=0.10\linewidth]
{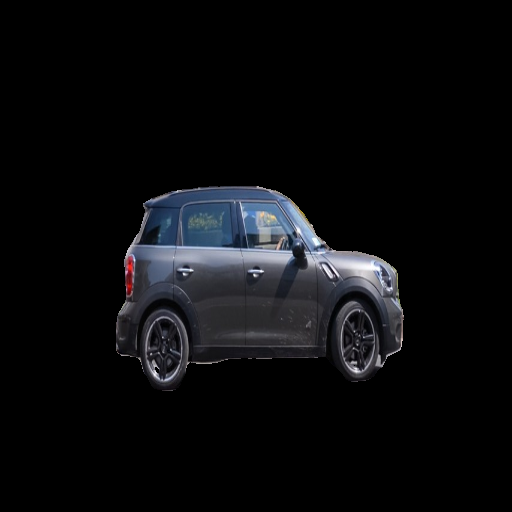} \ &
\includegraphics[width=0.15\linewidth, height=0.10\linewidth]
{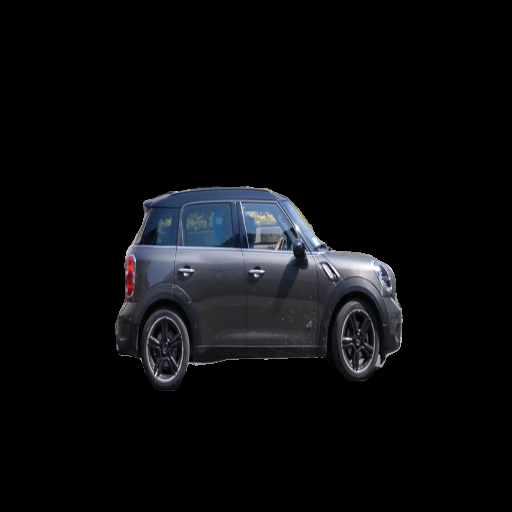} \\
\rotatebox{90}{\centering{\small{CRN}}}  \ &
\includegraphics[width=0.15\linewidth, height=0.10\linewidth]
{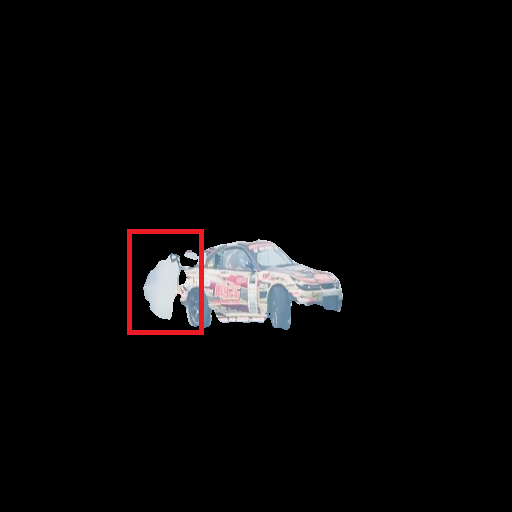} \ &
\includegraphics[width=0.15\linewidth, height=0.10\linewidth]
{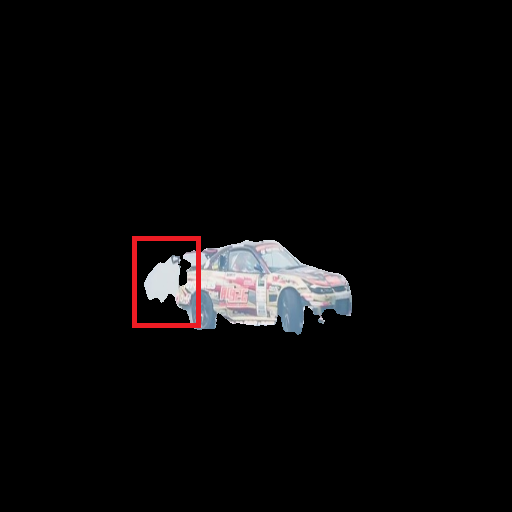} \ &
\includegraphics[width=0.15\linewidth, height=0.10\linewidth]
{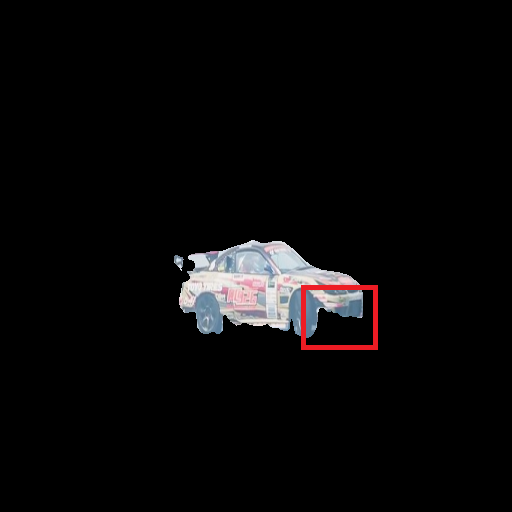} \ &
\includegraphics[width=0.15\linewidth, height=0.10\linewidth]
{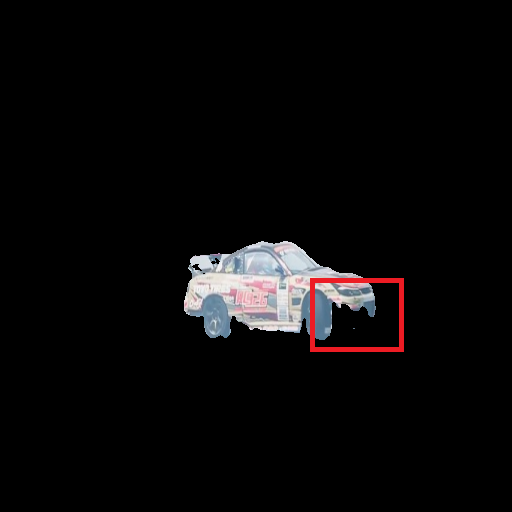} \ &
\includegraphics[width=0.15\linewidth, height=0.10\linewidth]
{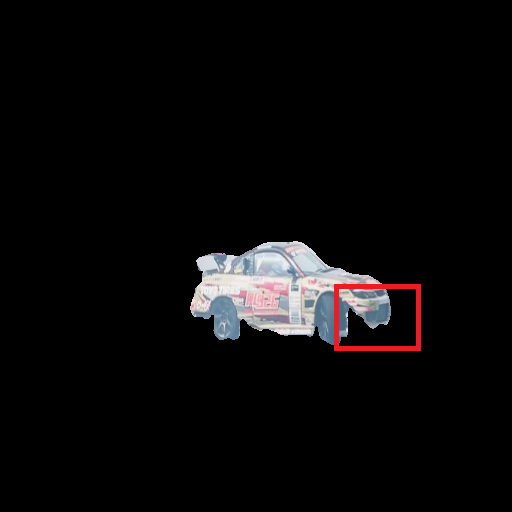} \ &
\includegraphics[width=0.15\linewidth, height=0.10\linewidth]
{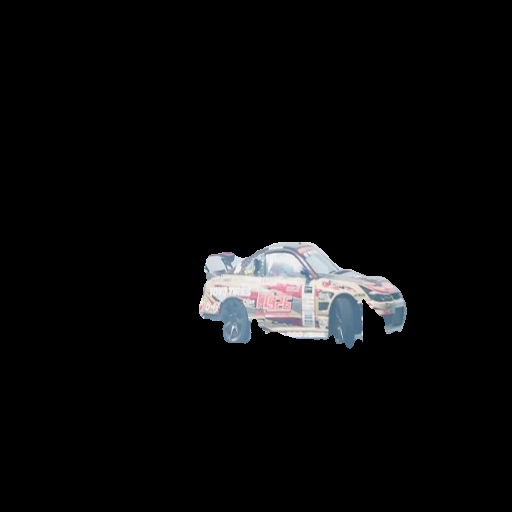} \\
\rotatebox{90}{\centering{\small{CRN+CL}}}  \ &
\includegraphics[width=0.15\linewidth, height=0.10\linewidth]
{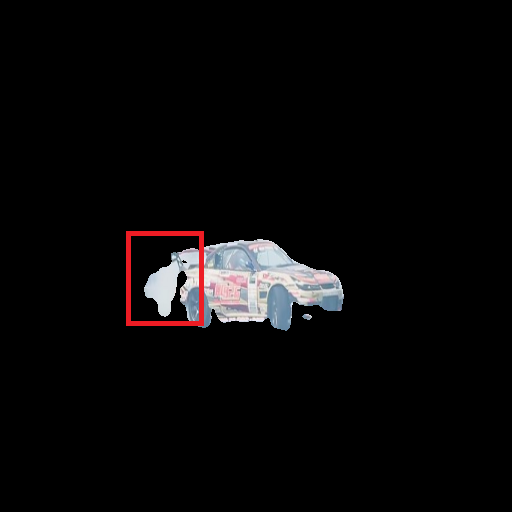} \ &
\includegraphics[width=0.15\linewidth, height=0.10\linewidth]
{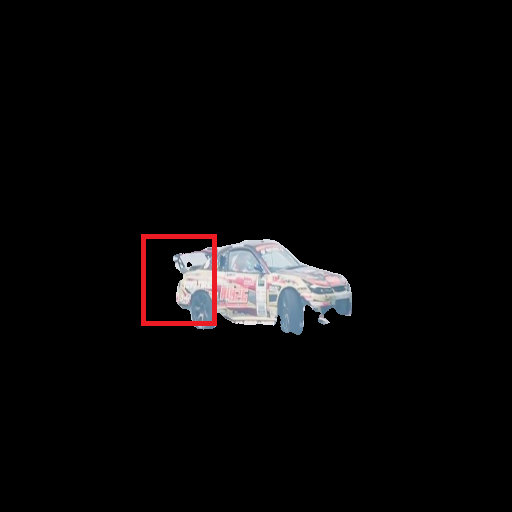} \ &
\includegraphics[width=0.15\linewidth, height=0.10\linewidth]
{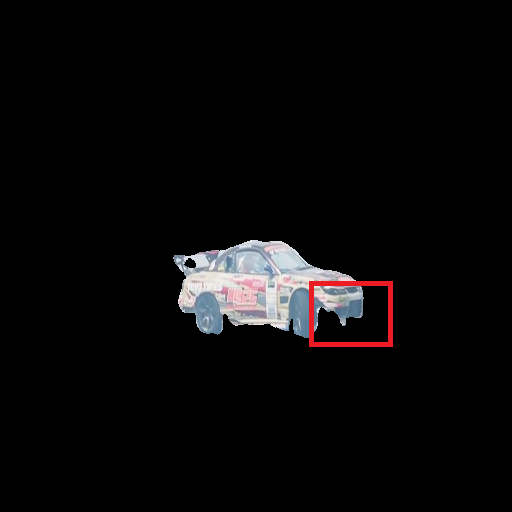} \ &
\includegraphics[width=0.15\linewidth, height=0.10\linewidth]
{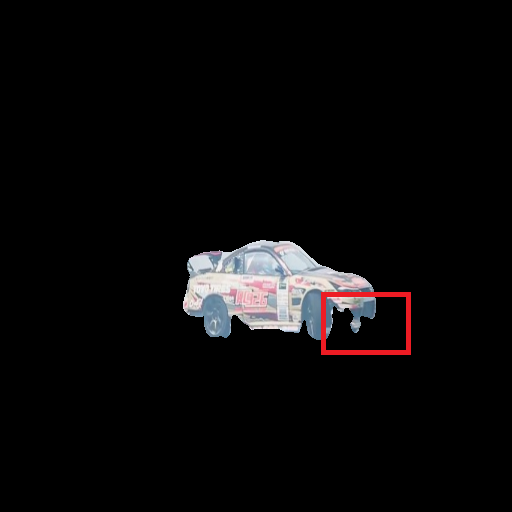} \ &
\includegraphics[width=0.15\linewidth, height=0.10\linewidth]
{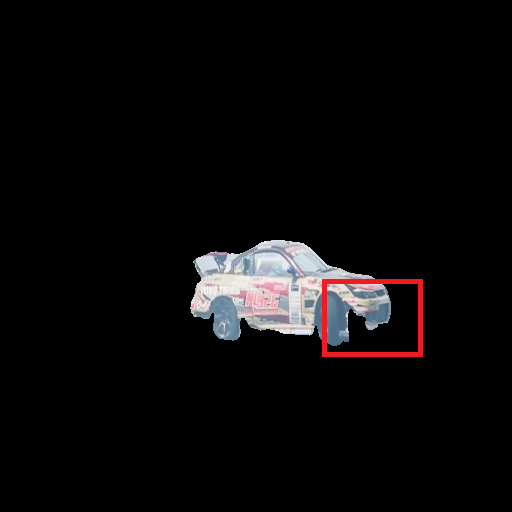} \ &
\includegraphics[width=0.15\linewidth, height=0.10\linewidth]
{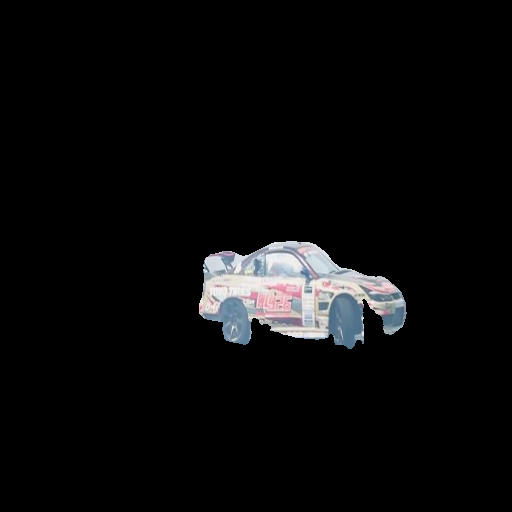} \\
\rotatebox{90}{\centering{\small{CRN}}}  \ &
\includegraphics[width=0.15\linewidth, height=0.10\linewidth]
{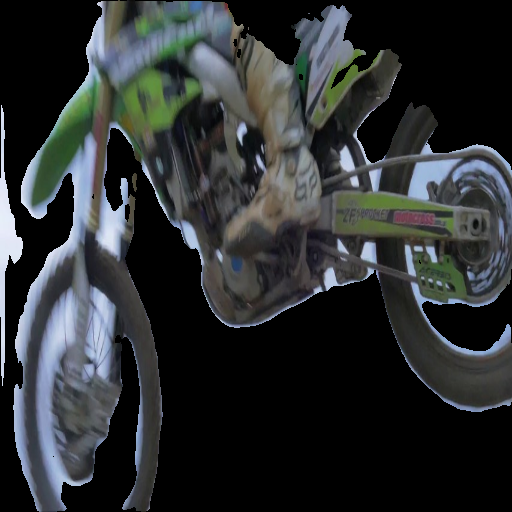} \ &
\includegraphics[width=0.15\linewidth, height=0.10\linewidth]
{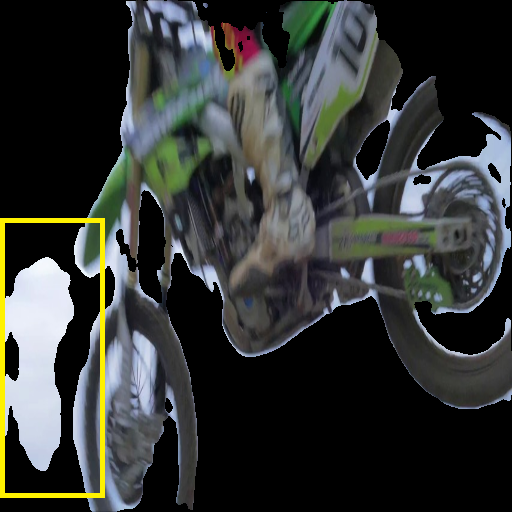} \ &
\includegraphics[width=0.15\linewidth, height=0.10\linewidth]
{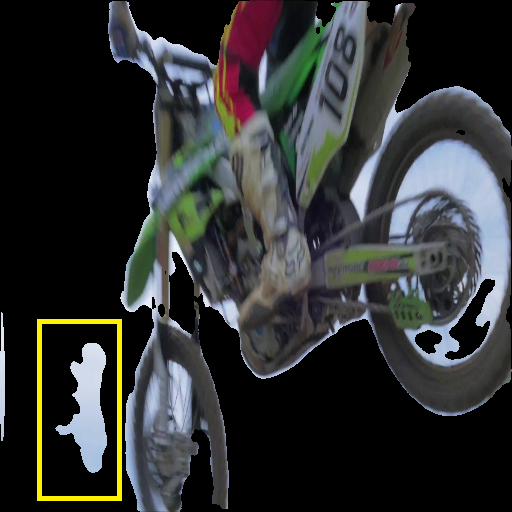} \ &
\includegraphics[width=0.15\linewidth, height=0.10\linewidth]
{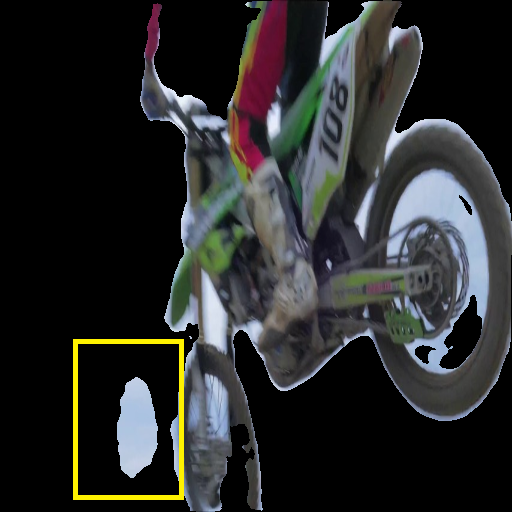} \ &
\includegraphics[width=0.15\linewidth, height=0.10\linewidth]
{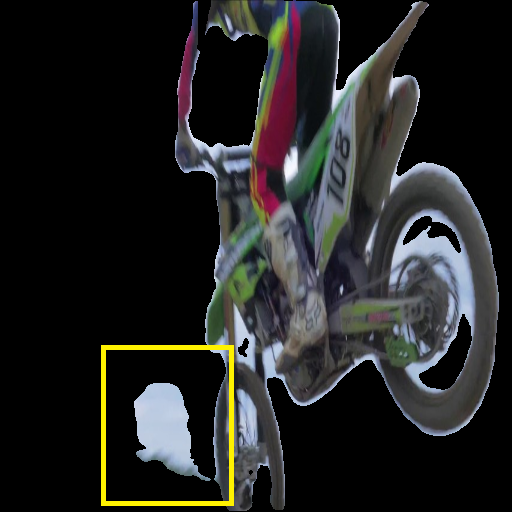} \ &
\includegraphics[width=0.15\linewidth, height=0.10\linewidth]
{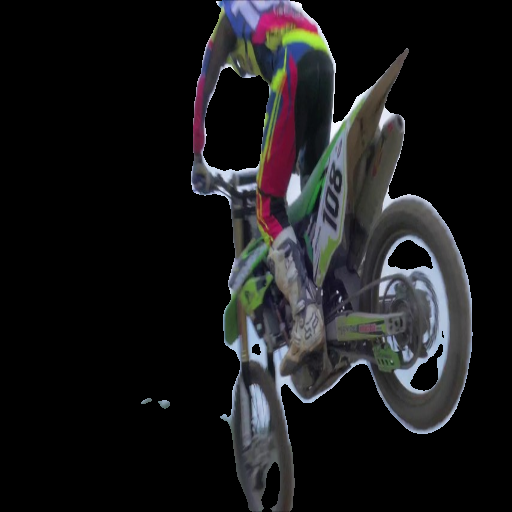} \\
\rotatebox{90}{\centering{\small{CRN+CL}}}  \ &
\includegraphics[width=0.15\linewidth, height=0.10\linewidth]
{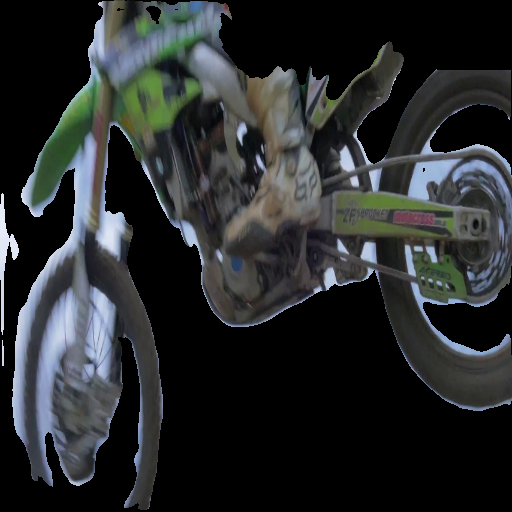} \ &
\includegraphics[width=0.15\linewidth, height=0.10\linewidth]
{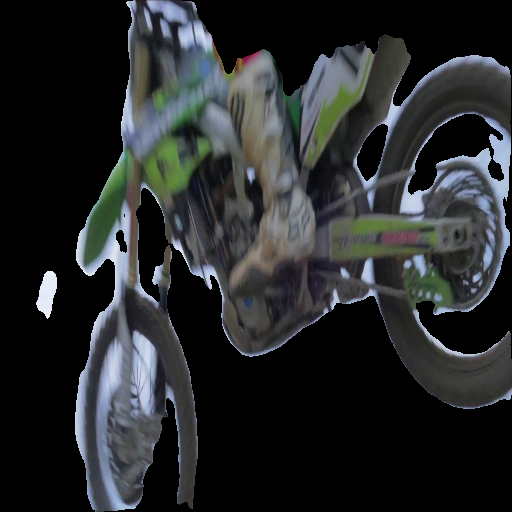} \ &
\includegraphics[width=0.15\linewidth, height=0.10\linewidth]
{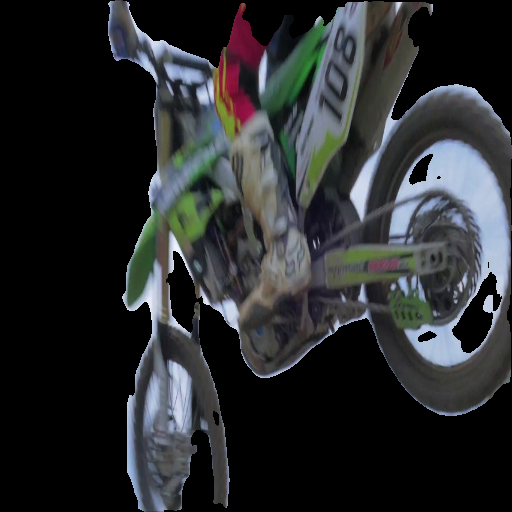} \ &
\includegraphics[width=0.15\linewidth, height=0.10\linewidth]
{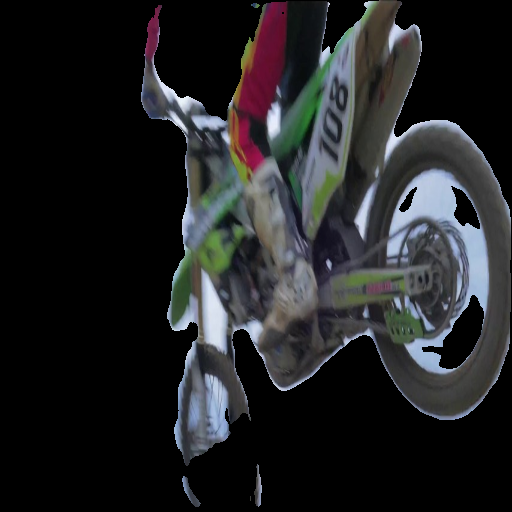} \ &
\includegraphics[width=0.15\linewidth, height=0.10\linewidth]
{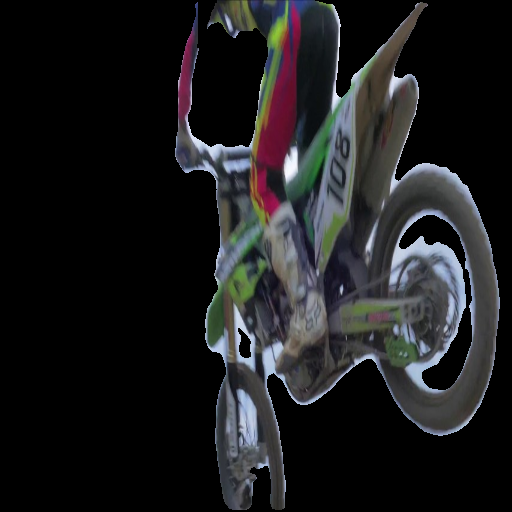} \ &
\includegraphics[width=0.15\linewidth, height=0.10\linewidth]
{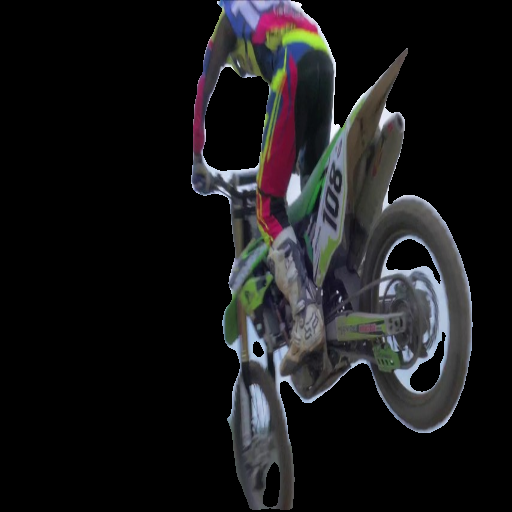}  \\
& frame $t$ & frame $t+1$ &frame $t+2$ &frame $t+3$ &frame $t+4$ & frame $t+5$
\end{tabular}
\end{center}
\caption{Qualitative video results for comparison of the CRN with/without CL on DAVIS, the rest is the results on DAVIS. We show significant improvement for stable results across adjacent frames.}
\label{Fig_dav}
\end{figure*}

\textbf{Results.}
As shown in Table~\ref{table:vhd}, in terms of the baseline SCHP trained with different backbones, our Coherent Loss(+CL) show great improvements on $STB$, especially along the object boundary, where our CL improves the SCHP(ResNet50) by 1.4\%.
Both the Boundary Coherency(+bc) and Global Coherency(+gc) improve the baseline on $STB$ and mIoU on synthetic set, which demonstrates that our Coherent Loss behaves effectively for enhancing the temporal coherency.
More importantly, from Figure~\ref{Fig_results} we show that with the Coherent Loss, the misalignment along the boundary is significantly improved and the visual performance become more smooth and satisfactory, and mis-segmentation are suppressed in the end.
Videos in the Supplementary Materials show more intuitive results.

\subsection{Video Object Segmentation}
\textbf{Dataset.}
We also evaluate our method on the public dataset: DAVIS-2016, which contains 50 video sequences in total, with 30 in the train set and 20 in the val set, and provides binary segmentation ground truth masks for all 3455 frames.

\textbf{Experiment Settings.}
We separately use the adjusted network SCHP and CRN~\cite{hu2018motion} as our baseline on which we further perform our Coherent Loss. CRN is a also standard network that takes the coarse segmentation as guidance to generate an accurate segmentation on DAVIS.

At the first step, for SCHP we use the similar training strategy on VHS for the baseline training on the training set, and for CRN we use official checkpoint model as the baseline.
When performing the Coherent Loss, we first make sure the segmentation loss has converged before activating the coherent supervision.
At the second step, we directly use the two consecutive frames from the training set as a image pair.
The fine-tune strategy of SCHP is similar to VHS, and for CRN we follow the original settings described in~\cite{hu2018motion} but the ratio of labeled images and unlabeled pairs for each batch are balanced.
We fine-tune the SCHP and CRN for 10 epoches combined with Coherent Loss.
The width $\theta$ is set to 15 and the threshold $\gamma$ is 5e-2 according to our research.
The weight $\alpha$ and the $\beta$ is set to 0.1 and 5e-6 for CRN, and 1 and 5e-5 for SCHP, enabling a balance between the supervised loss and unsupervised loss.

The metrics of the evaluation are the mIoU on the DAVIS-2016 val set, ${STB}_{global}$ and ${STB}_{local}$ on the synthetic test set, which is collected by using the first frame per video on DAVIS val set and is totally $20\times6\times11$ images.

\textbf{Results.}
As shown in Table~\ref{table:headings}, in terms of ${STB}$, our Coherent Loss both improves the CRN on global and boundary areas by 0.2\% and 0.7\%, which significantly achieve a promotion to produce stabler segmentation results, shown in Figure~\ref{Fig_dav}.
Although we focus on more coherent segmentation and better visual performance, the Coherent Loss, performing to exploits unlabeled videos to enhance temporal coherency, may also provide a valid temporal information for network to produce more accurate results.
Both CRN and SCHP are improved by Coherent Loss on mIoU and ${STB}$ with offline inference, thus this result demonstrates the flexibility and effectiveness that we can cooperate with many state-of-the-art segmentation methods based on neural network for better visual performance without adding extra prediction overhead.
We will show more results and the videos in the Supplementary Materials.

\subsection{Video Semantic Segmentation}
We also evaluate our method on Cityscape, which is a challenging dataset containing high quality pixel-level annotations for 5000 images.
The standard dataset split is 2975, 500, and 1525 for the training, validation, and test sets respectively.
Since the dataset is quite different from video object segmentation, we perform our Coherent Loss on the standard PSPNet and fine-tune it for 20 epoches with similar settings in~\cite{song2018pyramid}.
For the 3 provided unlabeled videos with successive frames, we use 2 videos for training and 1 for testing.
The width $\theta$ is set to 15 and the $\gamma$ is 0.3, the weight $\alpha$ and $\beta$ is set to 2e-2 and 0.3, enabling a balance between the supervised loss and unsupervised loss.
Eventually, we determine the best model in terms of mIoU on the whole val set and calculate the ${STB}$ on the synthetic set, in which the first frame is randomly chosen in the val set.

\begin{figure*}[tbp]
\begin{center}
\begin{tabular}{c@{}c@{}c@{}c@{}c@{}c@{}c@{}}\normalsize
\rotatebox{90}{\centering{\small{PSPNet}}} \ &
\includegraphics[width=0.15\linewidth, height=0.14\linewidth]
{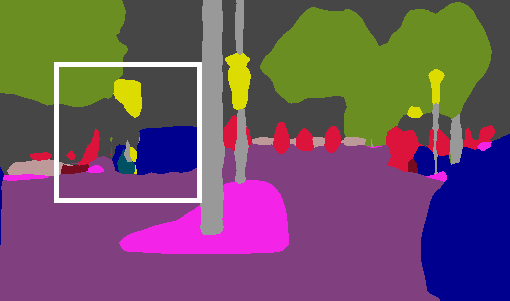} \ &
\includegraphics[width=0.15\linewidth, height=0.14\linewidth]
{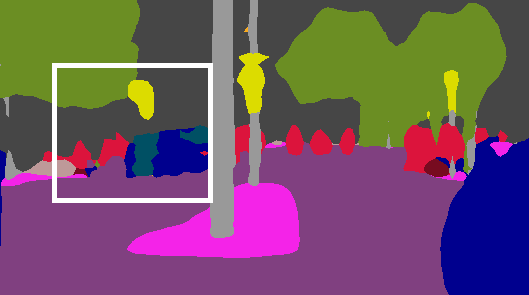} \ &
\includegraphics[width=0.15\linewidth, height=0.14\linewidth]
{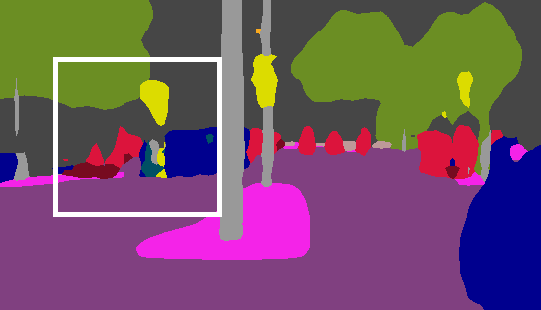} \ &
\includegraphics[width=0.15\linewidth, height=0.14\linewidth]
{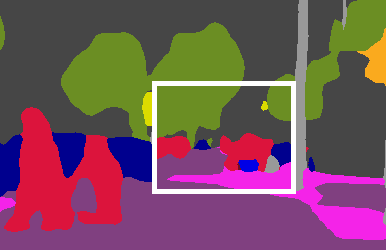} \ &
\includegraphics[width=0.15\linewidth, height=0.14\linewidth]
{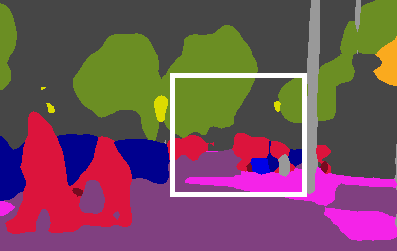} \ &
\includegraphics[width=0.15\linewidth, height=0.14\linewidth]
{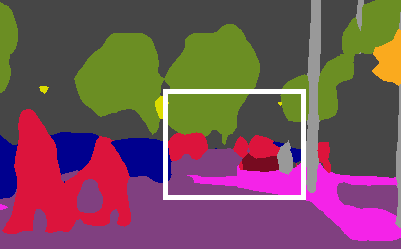} \\
\rotatebox{90}{\centering{\small{PSPNet+CL}}} \ &
\includegraphics[width=0.15\linewidth, height=0.14\linewidth]
{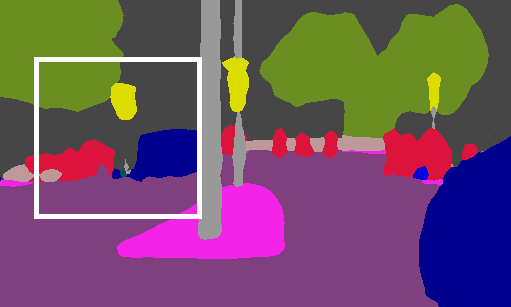} \ &
\includegraphics[width=0.15\linewidth, height=0.14\linewidth]
{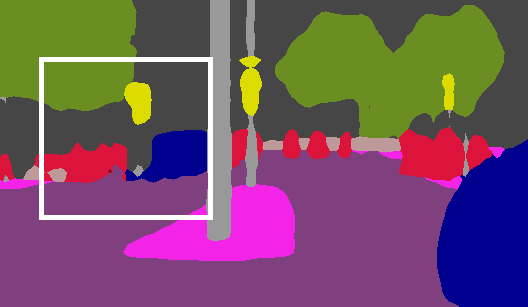} \ &
\includegraphics[width=0.15\linewidth, height=0.14\linewidth]
{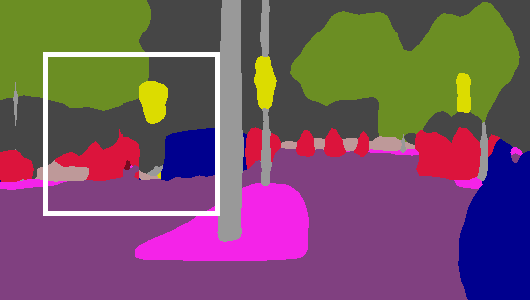} \ &
\includegraphics[width=0.15\linewidth, height=0.14\linewidth]
{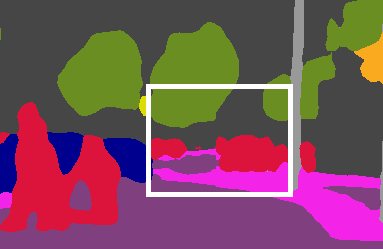}  \ &
\includegraphics[width=0.15\linewidth, height=0.14\linewidth]
{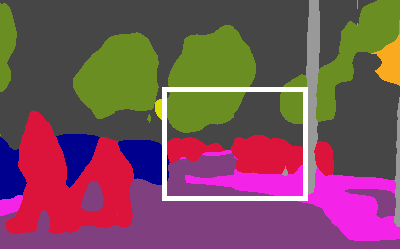}  \ &
\includegraphics[width=0.15\linewidth, height=0.14\linewidth]
{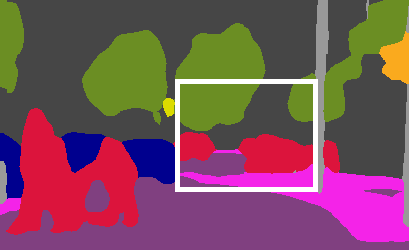} \\
& frame t & frame t+1 & frame t+2 & frame t & frame t+1  & frame t+2\\
\end{tabular}
\end{center}
\caption{Visual examples cropped from the original frames on Cityscapes. Our method further improves the misalignment along the boundary and reduces mis-segmentations.}
\label{Fig_psp}
\end{figure*}

We compare our method to the official baseline PSPNet with ResNet50 as a backbone in Table~\ref{table:pspnet}, and we achieve a significant improvement both on accuracy and stability.
and from the visual examples in Figure~\ref{Fig_psp}, we can see that although the baseline achieve high mIoU of the whole val set, many visual jitters are overlooked during the evaluation.
Equipped with Coherent Loss, the baseline achieve a significant improvement for stabler results.

\subsection{Discussion}
We carry out our method on three different video segmentation datasets to investigate its performance and the results show that our method significantly enhances the temporal coherency for all datasets, especially on the object boundary.
Visual performance in Figure~\ref{Fig_results} 
demonstrates that our method performs more stabler with fewer mis-segmentation and unaligned boundaries.
One may doubt that ${STB}$ may be very high when the network performs poorly in the whole sequences, we think it makes more sense to discuss stability on the basis of a certain segmentation accuracy.

In addition, our Coherent Loss is designed for enhancing the ability of a given baseline for providing satisfactory video segmentation results in term of temporal coherency, however, the results in Table~\ref{table:headings} demonstrates that our method can not only improve the ${STB}$, but also help to improve the segmentation accuracy in term of mIoU, which we believe is probably because more coherent results are more accurate especially when the object is notable in the video to segment.
Equipped with our method, many state-of-the-art segmentation approaches can achieve a significant improvement on more satisfactory visual quality on VHS and DAVIS.
However, our method may not bring a noticeable increase in accuracy on Cityscape, this is probably due to the fact that our Coherent Loss enforces the network to perform coherently for the whole image, thus those trembling small objects will be smoothed for stabler performance.

\begin{table}[tbp]
\begin{center}
\caption{Comparison of mIoU on Cityscape val set, ${STB}$ on the synthetic set.}
\label{table:pspnet}
\begin{tabular}{cccc}
\hline
 Method     & mIoU(val)   & ${STB}_{global}$   &${STB}_{local}$ \\
\hline
PSPNet      &  77.02      &   93.06            &    90.22       \\
PSPNet+CL   &  77.88      &   93.08            &    91.63       \\
\hline
\end{tabular}
\end{center}
\end{table}

\section{Conclusions}
Inspired by unstable phenomena in applications, we expose, temporal coherency, a new dimension of measuring the quality of video segmentation performance.
For explicitly evaluating, we hence introduce a new numeric metric, stability rate, which is verified by user-study.
Further more, we also propose Coherent Loss with a generic framework performing as a neural network training strategy to enhance the temporal coherency in video segmentation, which is capable of exploiting unlabeled videos to learn the boundary consistency and global consistency temporally.
Owing it is orthogonal to the network architecture, it hence can be used to cooperate with different segmentation models.
Comprehensive experiments are carried out on a variety of datasets, which verifies our method being an effective framework for providing better results in combination of segmentation accuracy and temporal coherency.
A large labeled video human segmentation dataset is collected for future study in this field and will be made available to the public.

\bibliographystyle{aaai}
\bibliography{egbib}

\newpage
\section{Dataset}
\begin{figure*}
\begin{center}
\begin{tabular}{c@{}c@{}c@{}c@{}c@{}c@{}c@{}}\small
\centering
\rotatebox{90}{\centering{ATR}} \ &
\includegraphics[width=0.15\linewidth, height=0.15\linewidth]
{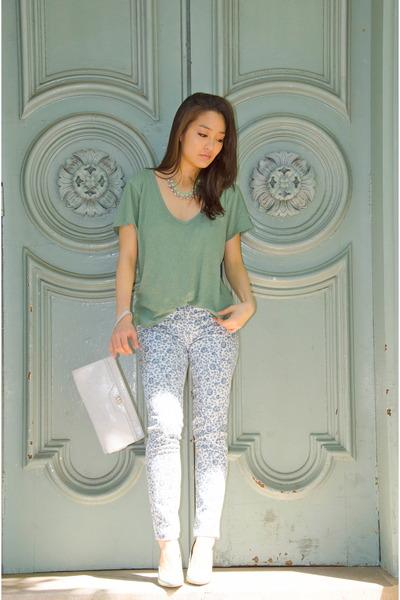} \ &
\includegraphics[width=0.15\linewidth, height=0.15\linewidth]
{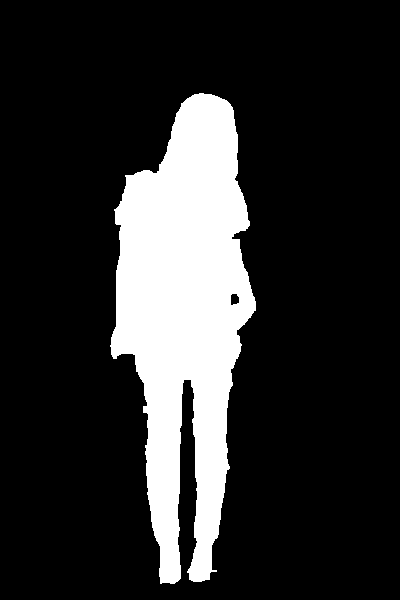} \ &
\includegraphics[width=0.15\linewidth, height=0.15\linewidth]
{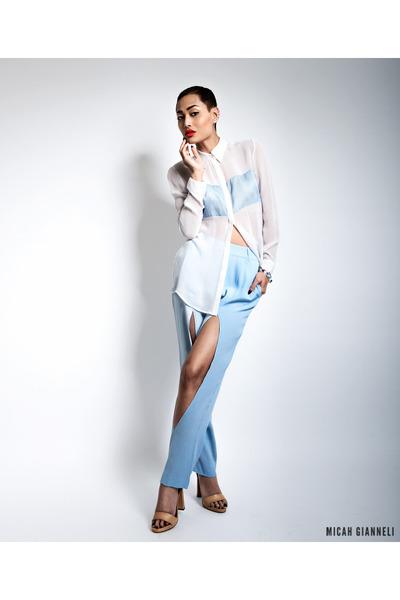} \ &
\includegraphics[width=0.15\linewidth, height=0.15\linewidth]
{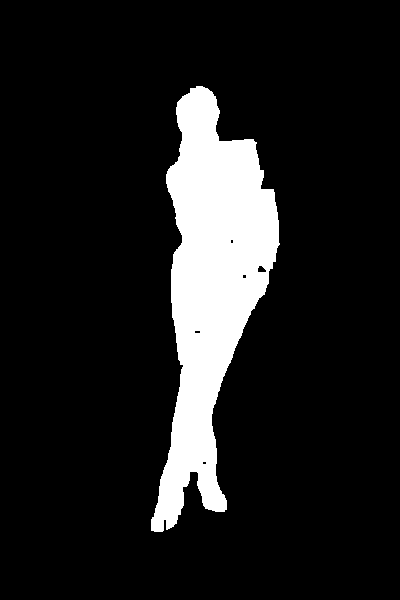}  \ &
\includegraphics[width=0.15\linewidth, height=0.15\linewidth]
{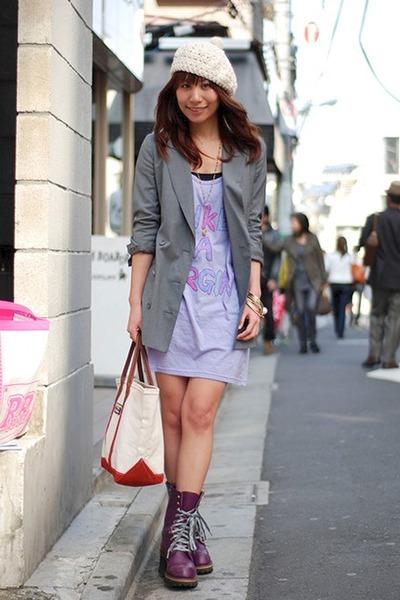} \ &
\includegraphics[width=0.15\linewidth, height=0.15\linewidth]
{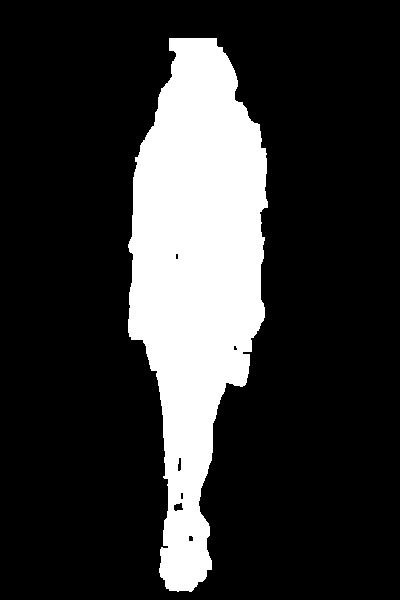} \\
\centering
\rotatebox{90}{\centering{AIC}} \ &
\includegraphics[width=0.15\linewidth, height=0.15\linewidth]
{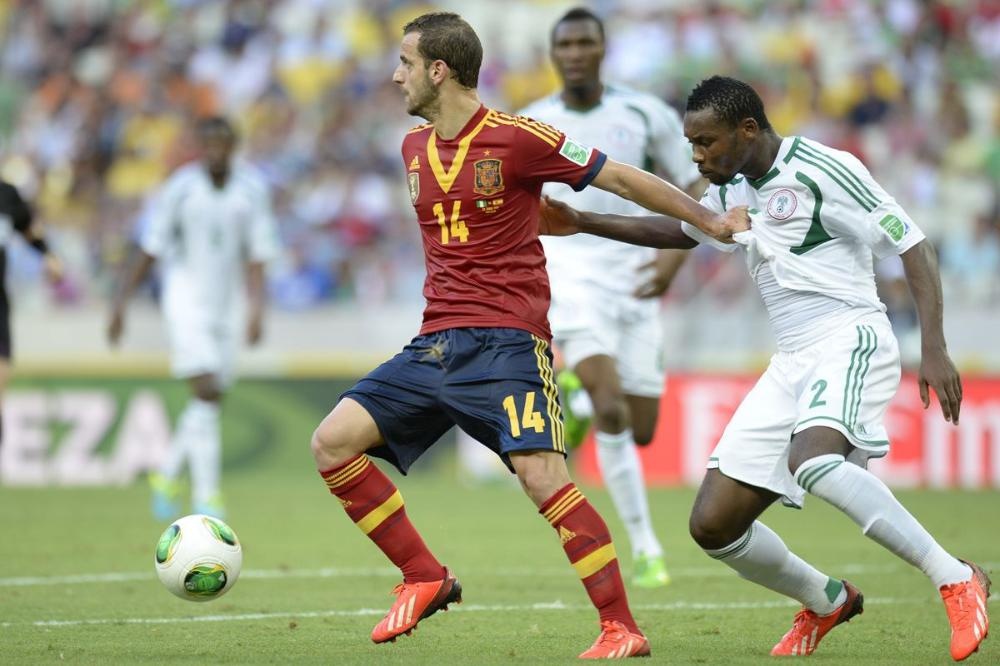} \ &
\includegraphics[width=0.15\linewidth, height=0.15\linewidth]
{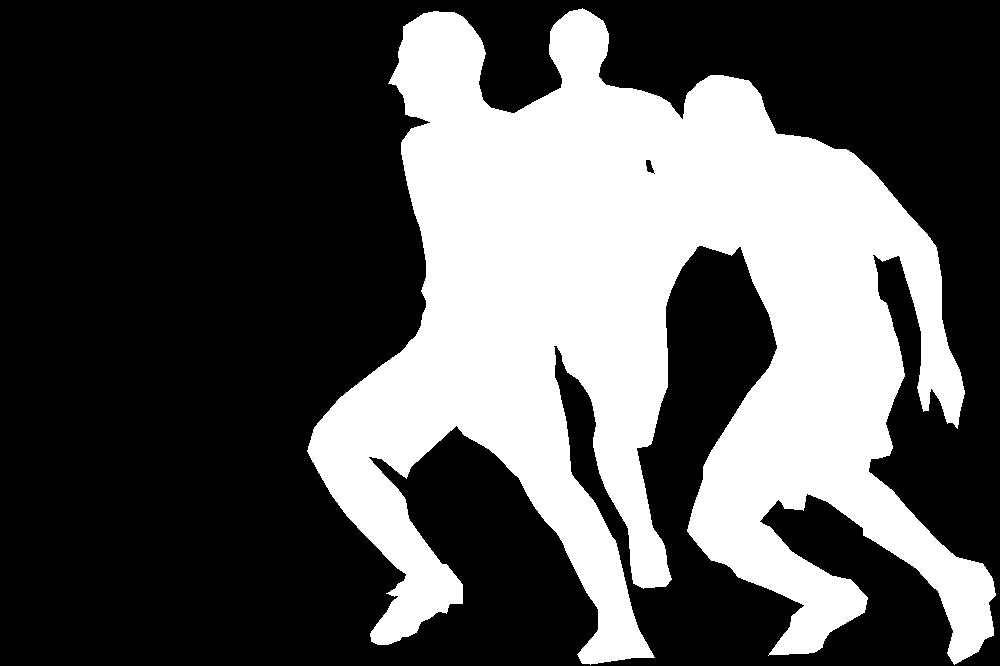} \ &
\includegraphics[width=0.15\linewidth, height=0.15\linewidth]
{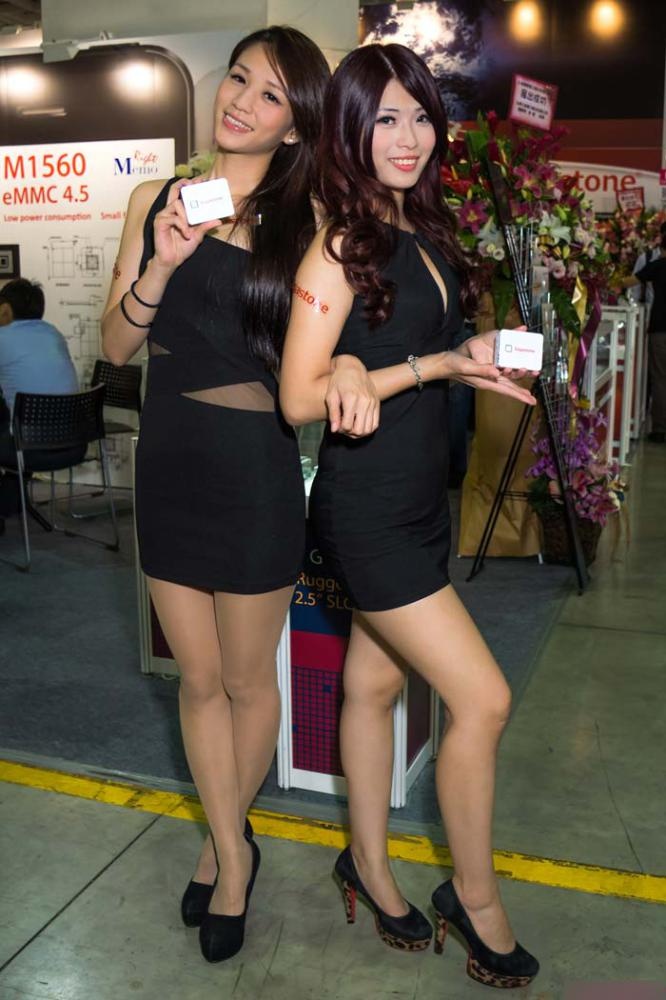} \ &
\includegraphics[width=0.15\linewidth, height=0.15\linewidth]
{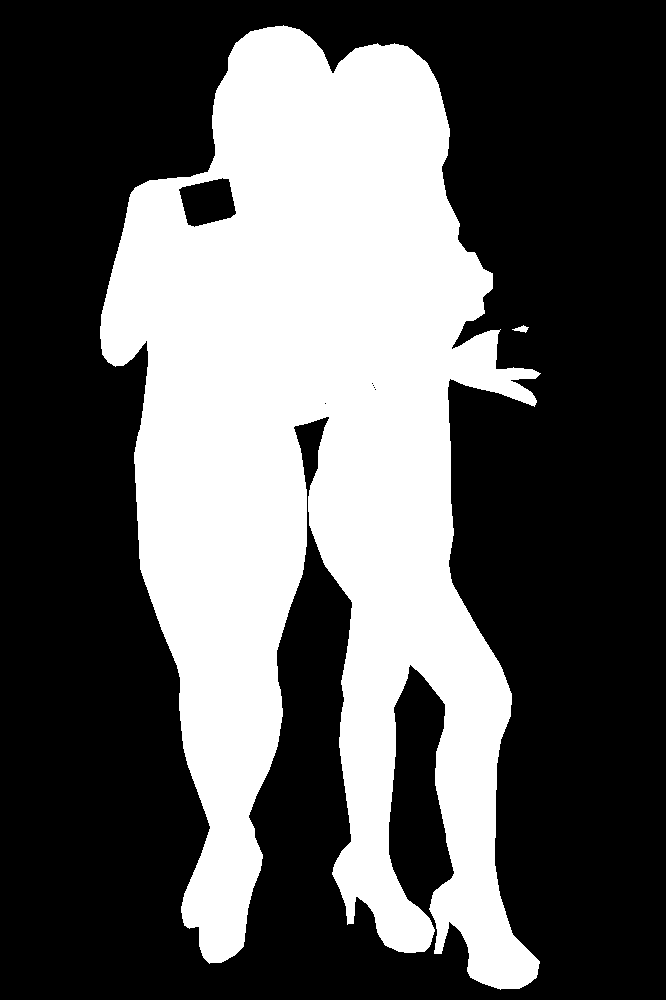}  \ &
\includegraphics[width=0.15\linewidth, height=0.15\linewidth]
{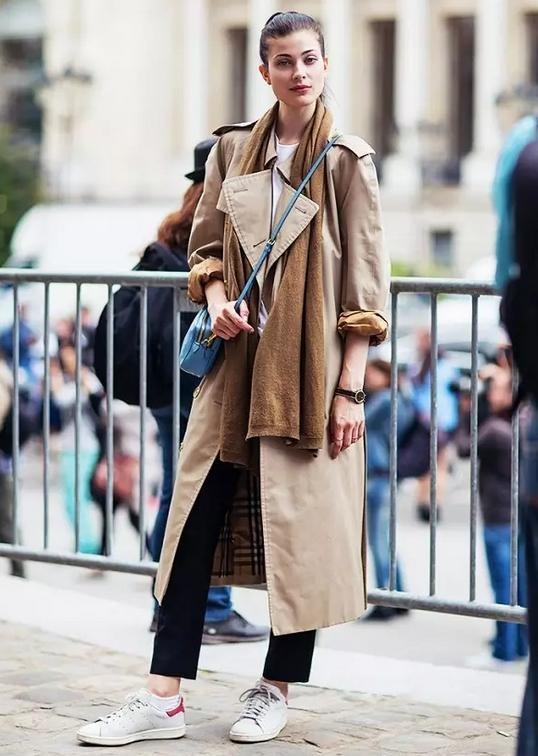} \ &
\includegraphics[width=0.15\linewidth, height=0.15\linewidth]
{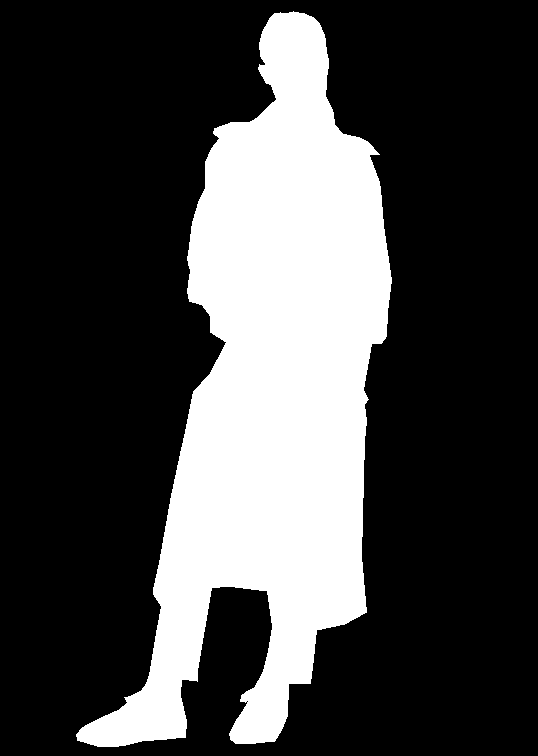} \\
\centering
\rotatebox{90}{\centering{AISegment}} \ &
\includegraphics[width=0.15\linewidth, height=0.15\linewidth]
{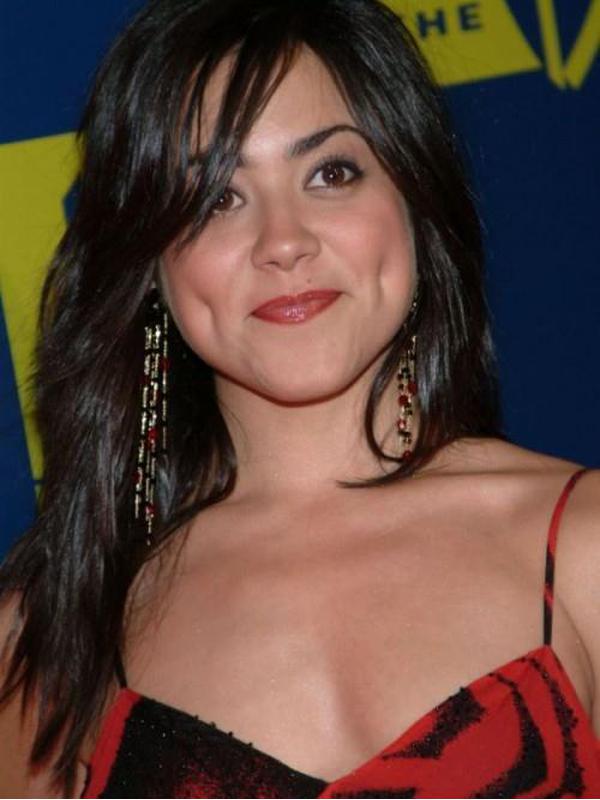} \ &
\includegraphics[width=0.15\linewidth, height=0.15\linewidth]
{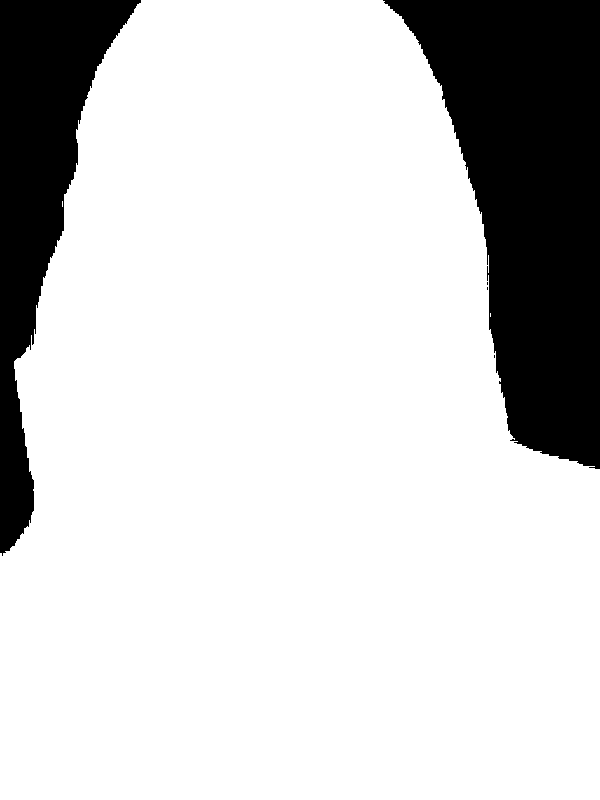} \ &
\includegraphics[width=0.15\linewidth, height=0.15\linewidth]
{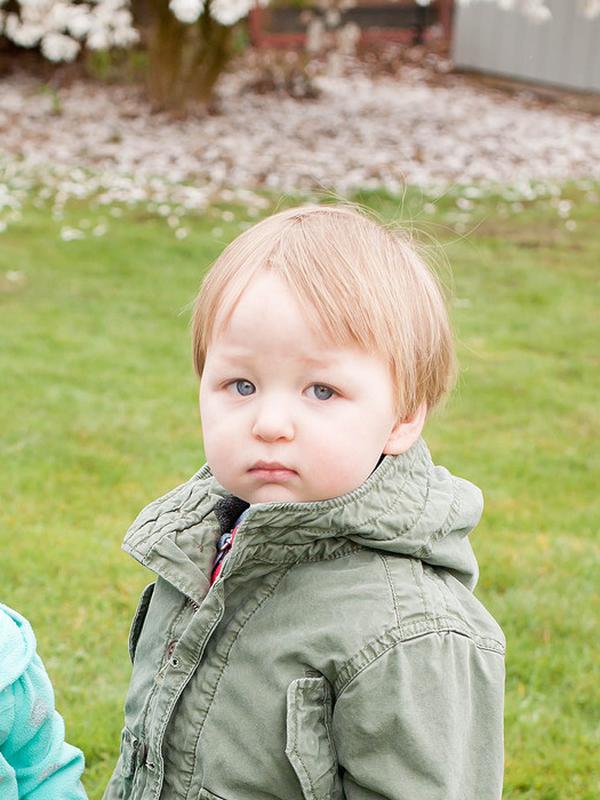} \ &
\includegraphics[width=0.15\linewidth, height=0.15\linewidth]
{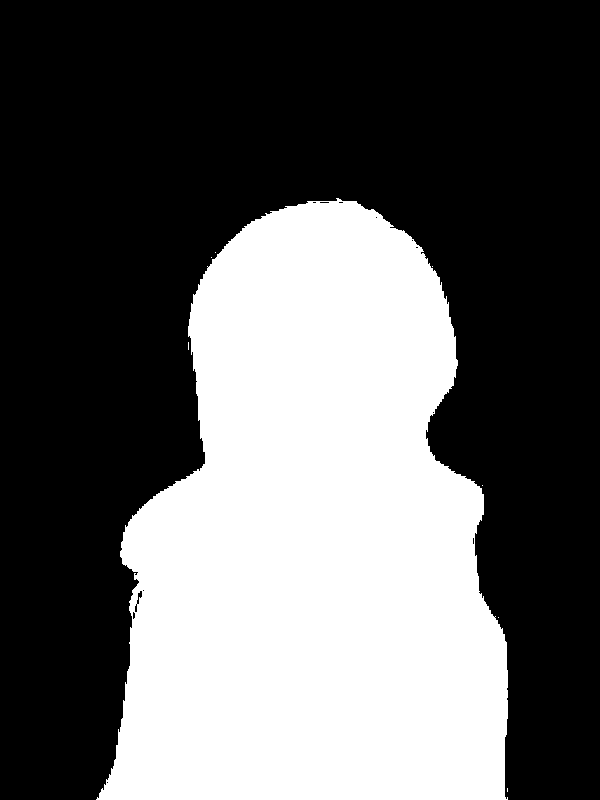}  \ &
\includegraphics[width=0.15\linewidth, height=0.15\linewidth]
{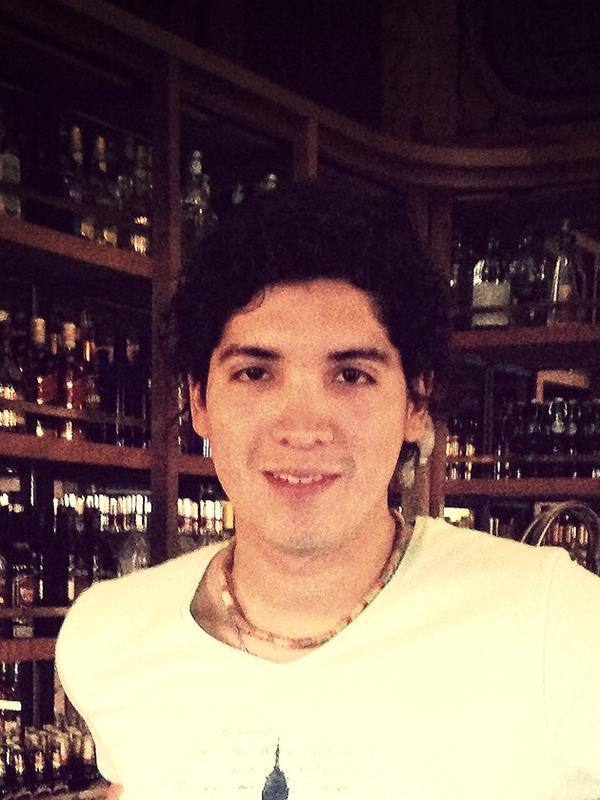} \ &
\includegraphics[width=0.15\linewidth, height=0.15\linewidth]
{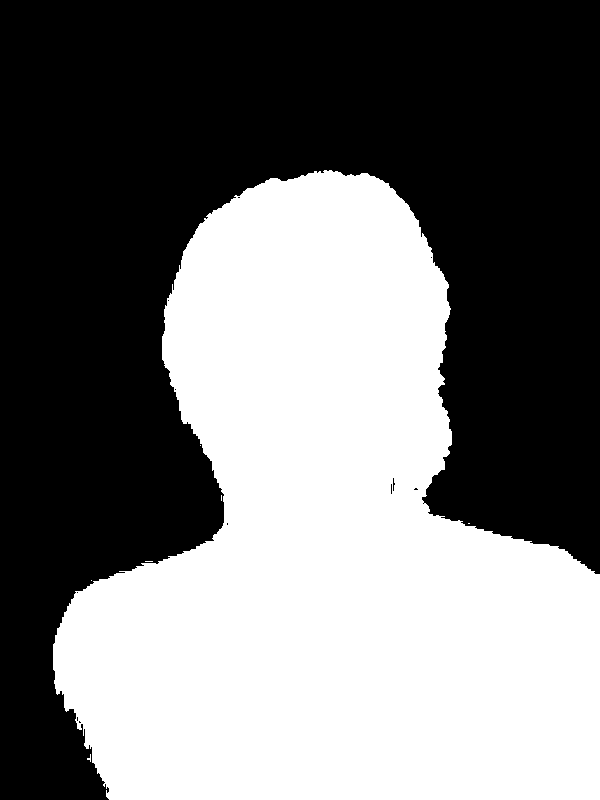} \\

\centering
\rotatebox{90}{\centering{LIP}} \ &
\includegraphics[width=0.15\linewidth, height=0.15\linewidth]
{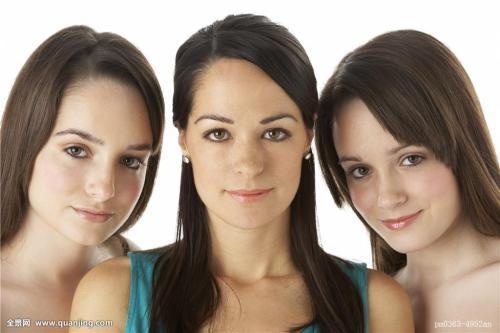} \ &
\includegraphics[width=0.15\linewidth, height=0.15\linewidth]
{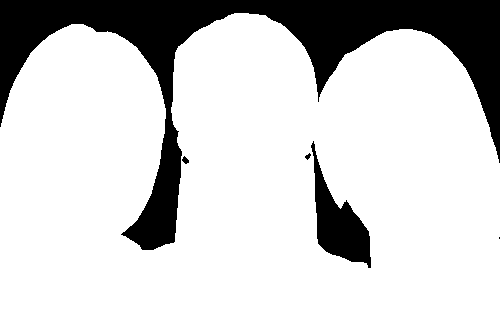} \ &
\includegraphics[width=0.15\linewidth, height=0.15\linewidth]
{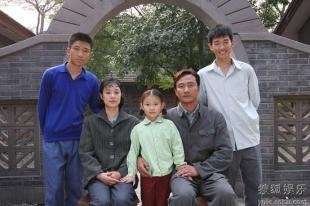} \ &
\includegraphics[width=0.15\linewidth, height=0.15\linewidth]
{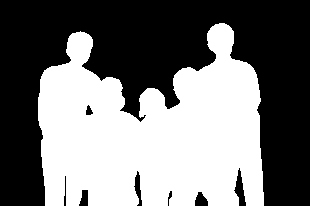}  \ &
\includegraphics[width=0.15\linewidth, height=0.15\linewidth]
{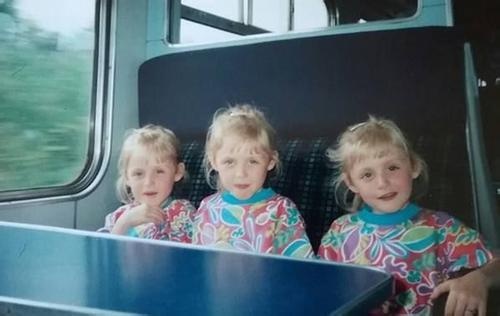} \ &
\includegraphics[width=0.15\linewidth, height=0.15\linewidth]
{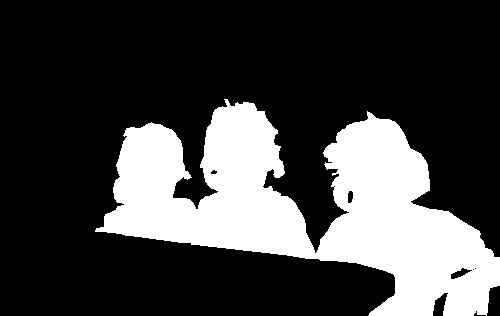} \\

\centering
\rotatebox{90}{\centering{SPD}} \ &
\includegraphics[width=0.15\linewidth, height=0.15\linewidth]
{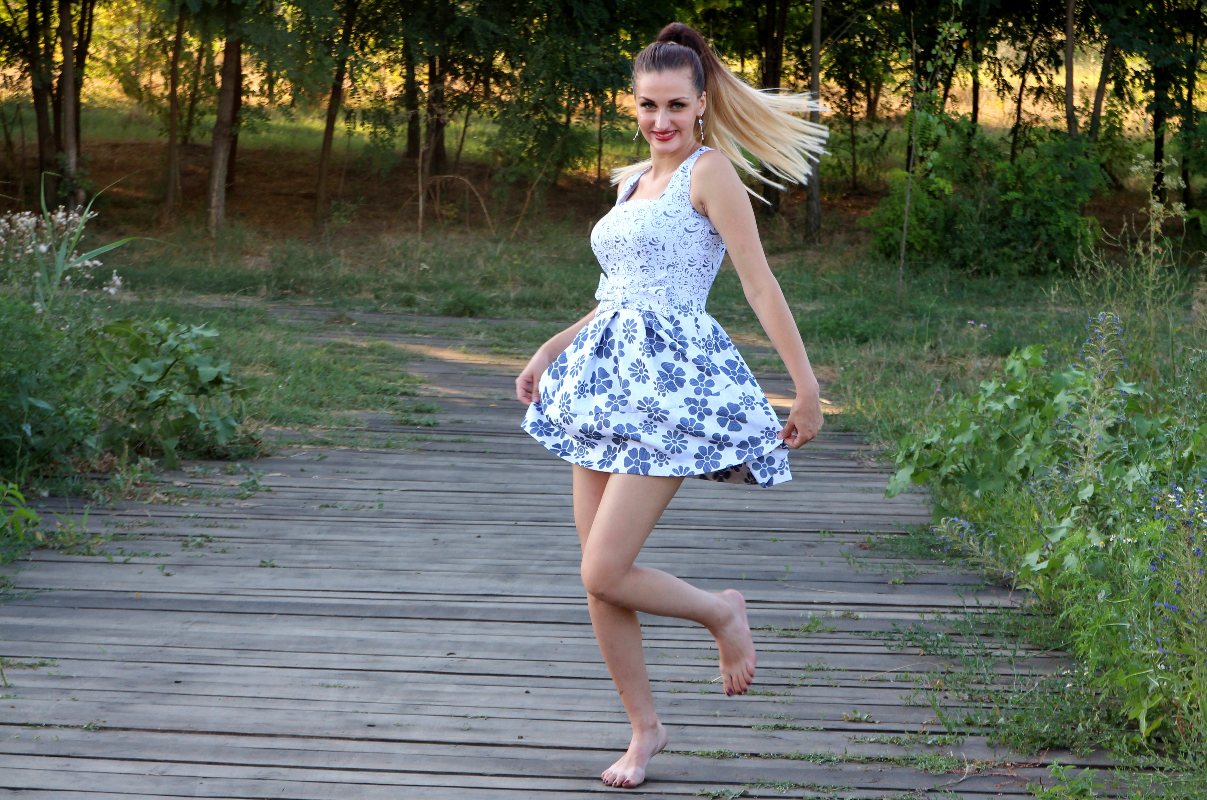} \ &
\includegraphics[width=0.15\linewidth, height=0.15\linewidth]
{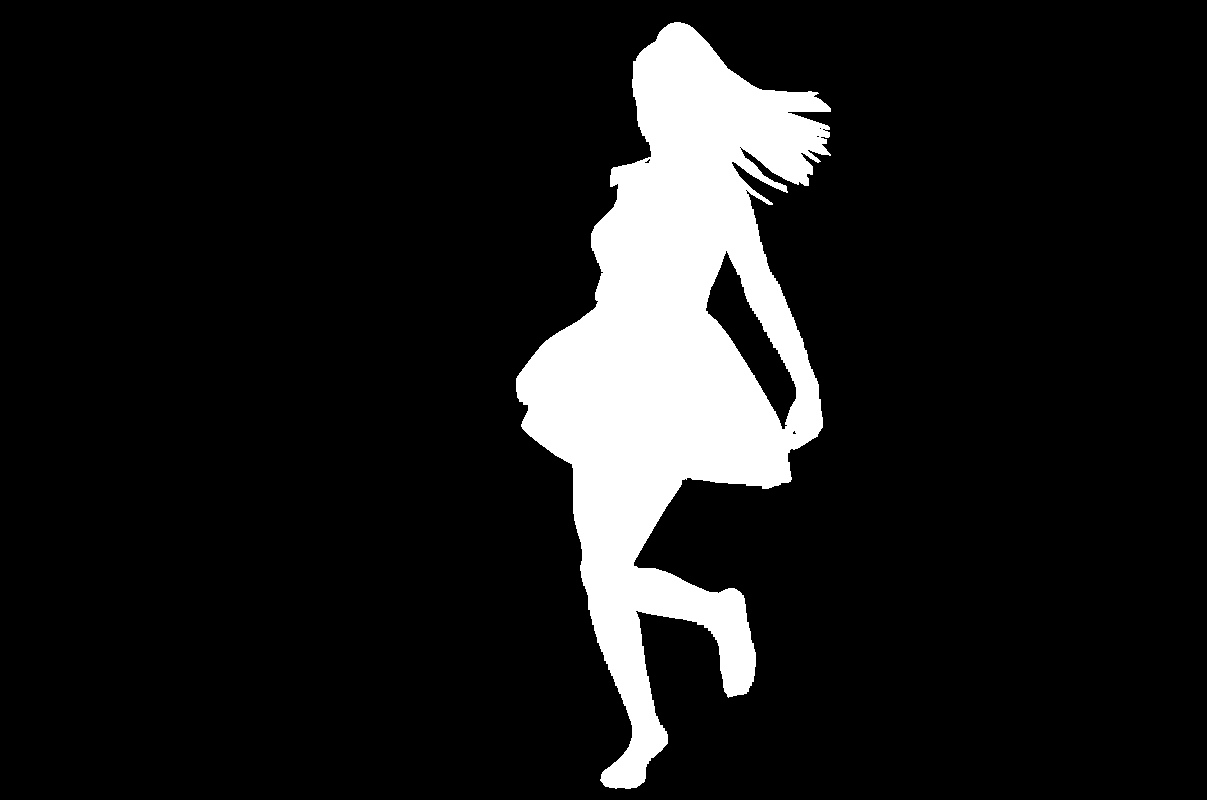} \ &
\includegraphics[width=0.15\linewidth, height=0.15\linewidth]
{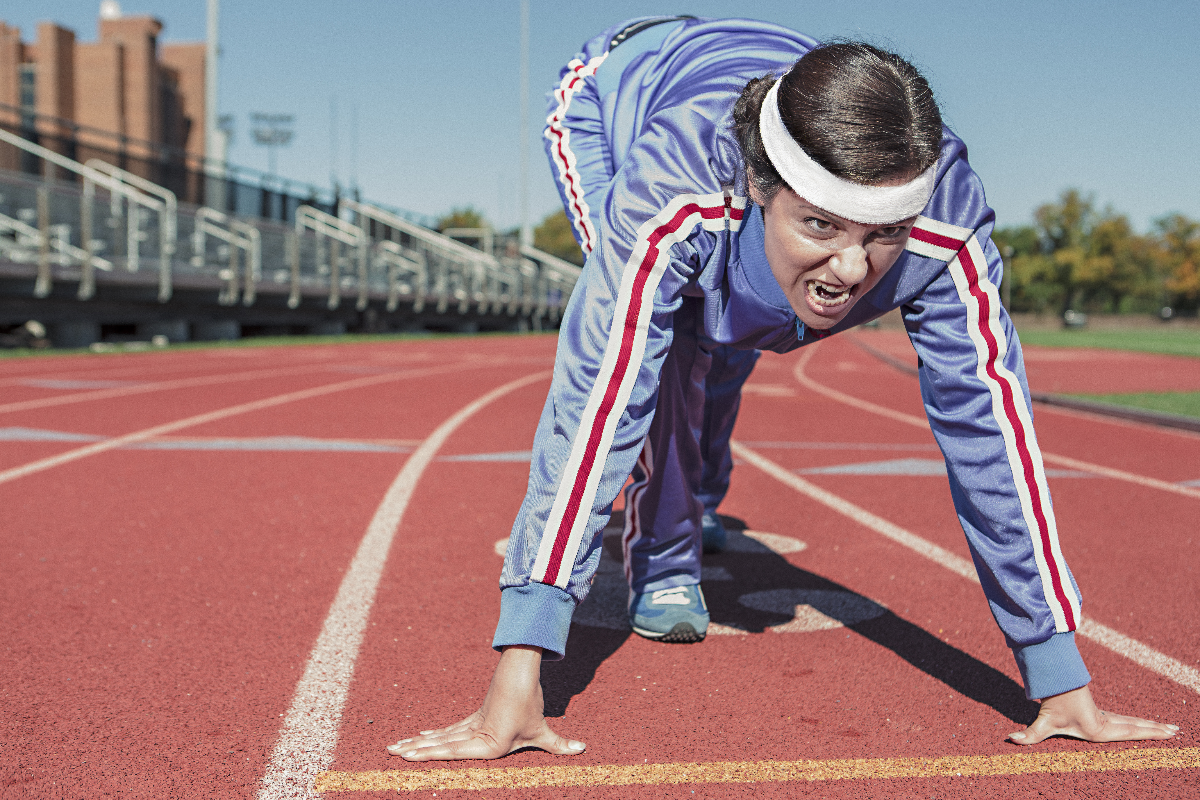} \ &
\includegraphics[width=0.15\linewidth, height=0.15\linewidth]
{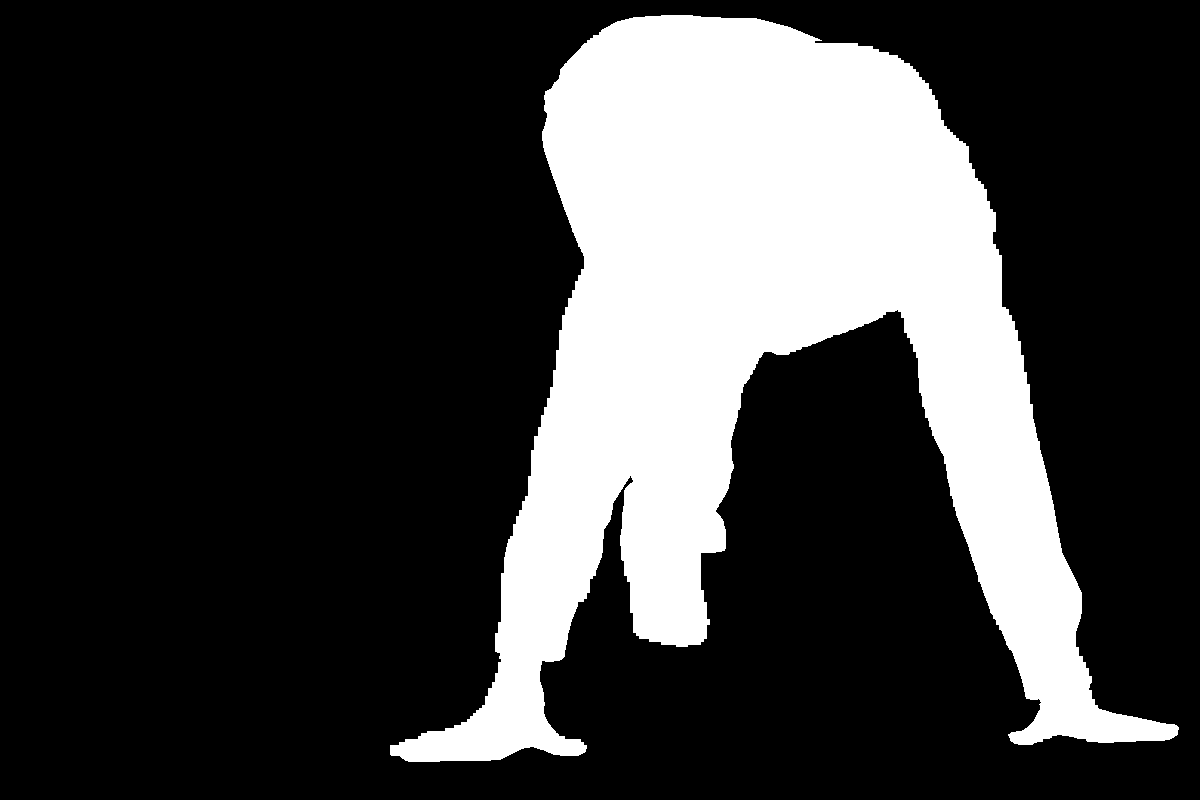} \ &
\includegraphics[width=0.15\linewidth, height=0.15\linewidth]
{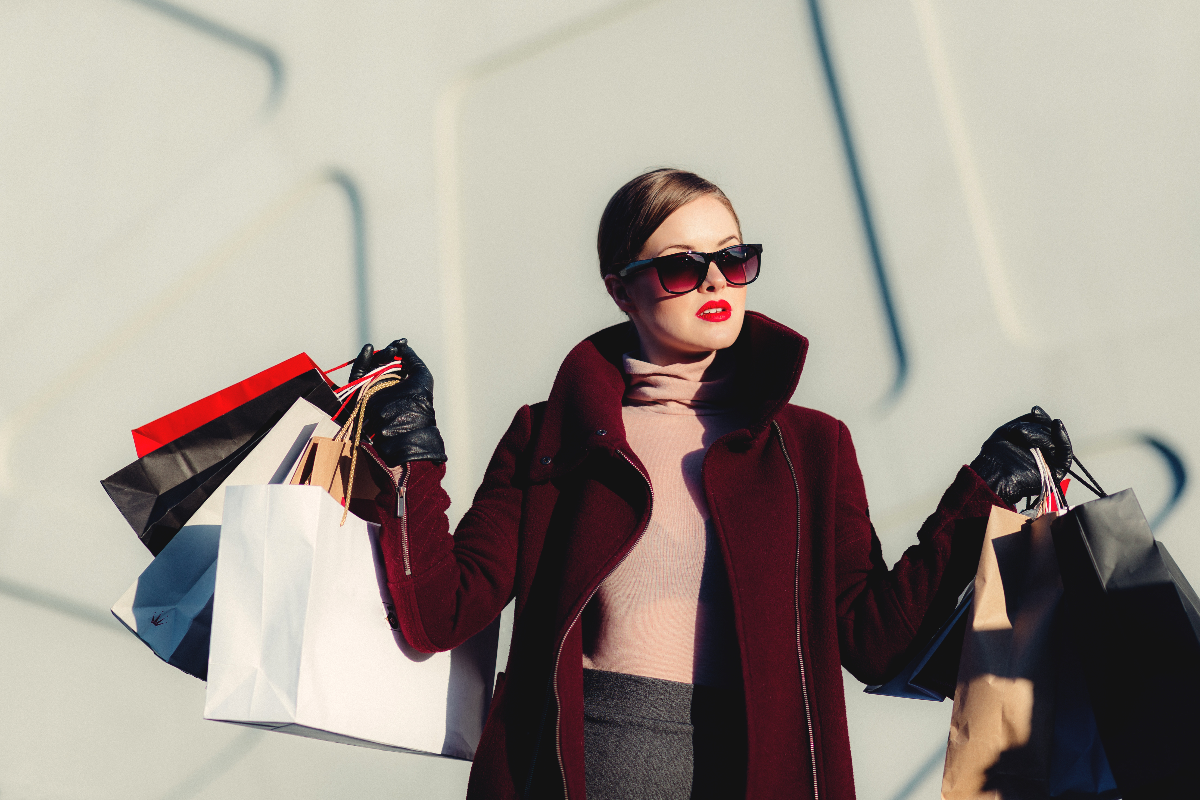} \ &
\includegraphics[width=0.15\linewidth, height=0.15\linewidth]
{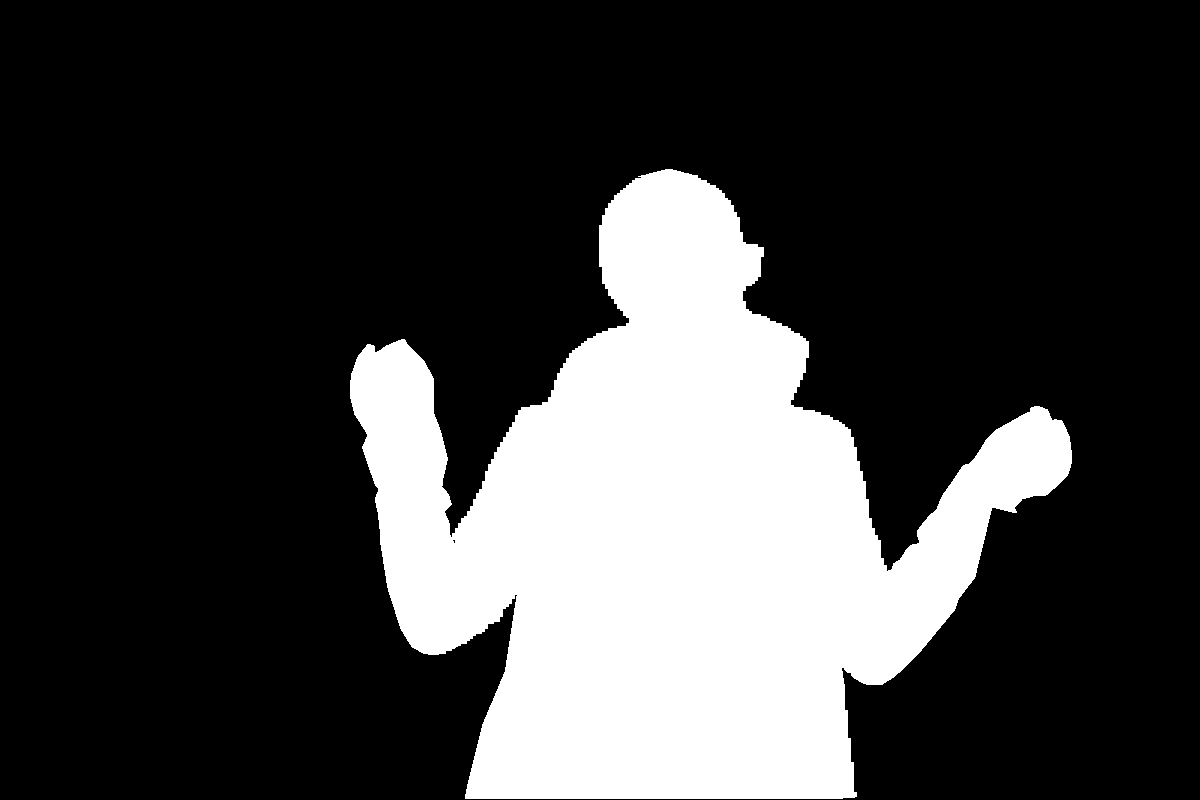} \\

& Image $i$ & label $i$ &Image $j$ &label $j$ &Image $k$ & label $k$
\end{tabular}
\end{center}
\caption{Images and labels on the video human segmentation.}
\label{Fig_results}
\end{figure*}

\begin{figure*}
\begin{center}
\begin{tabular}{c@{}c@{}c@{}c@{}c@{}c@{}}
\includegraphics[width=0.15\linewidth, height=0.15\linewidth]
{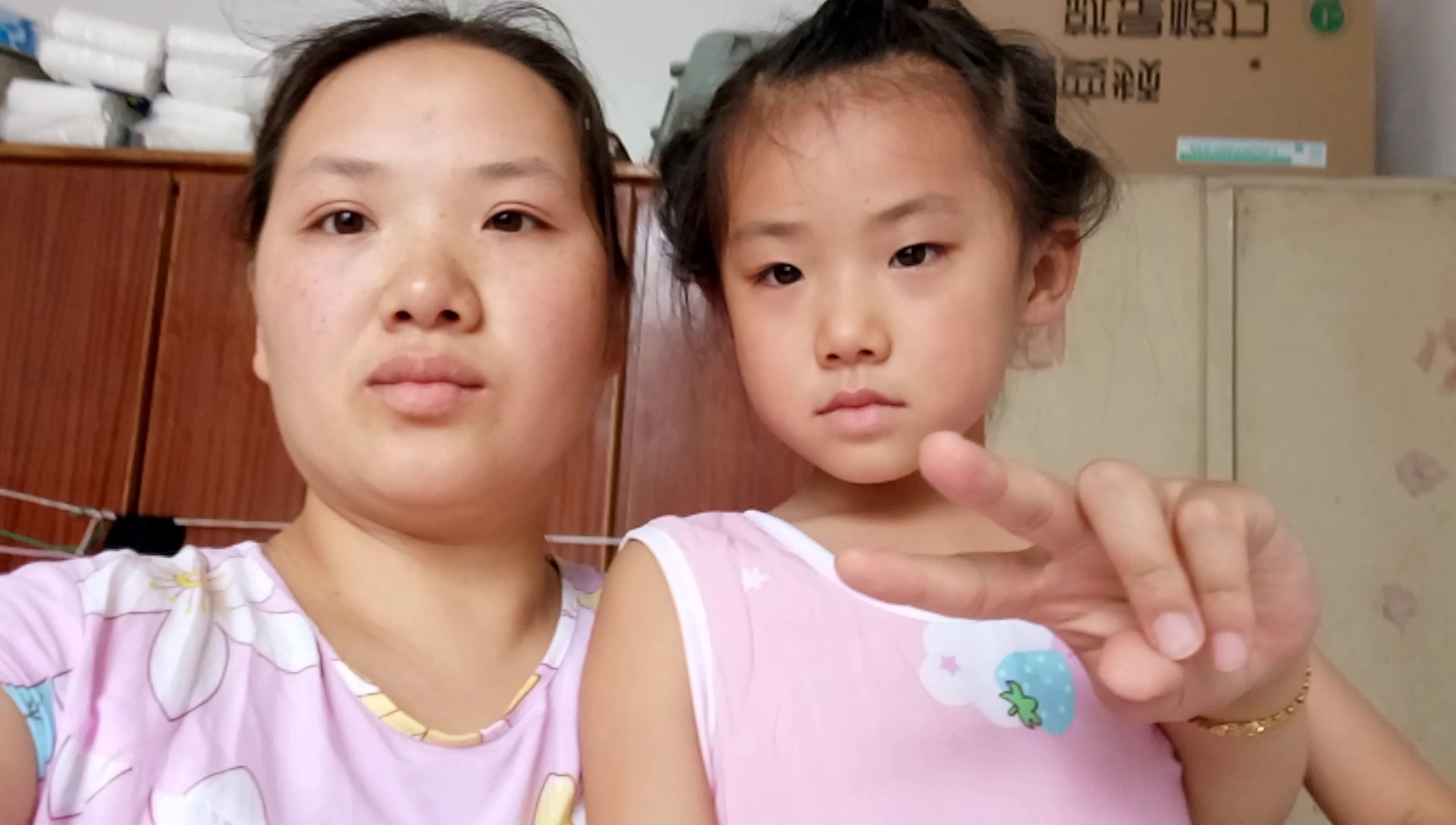} \ &
\includegraphics[width=0.15\linewidth, height=0.15\linewidth]
{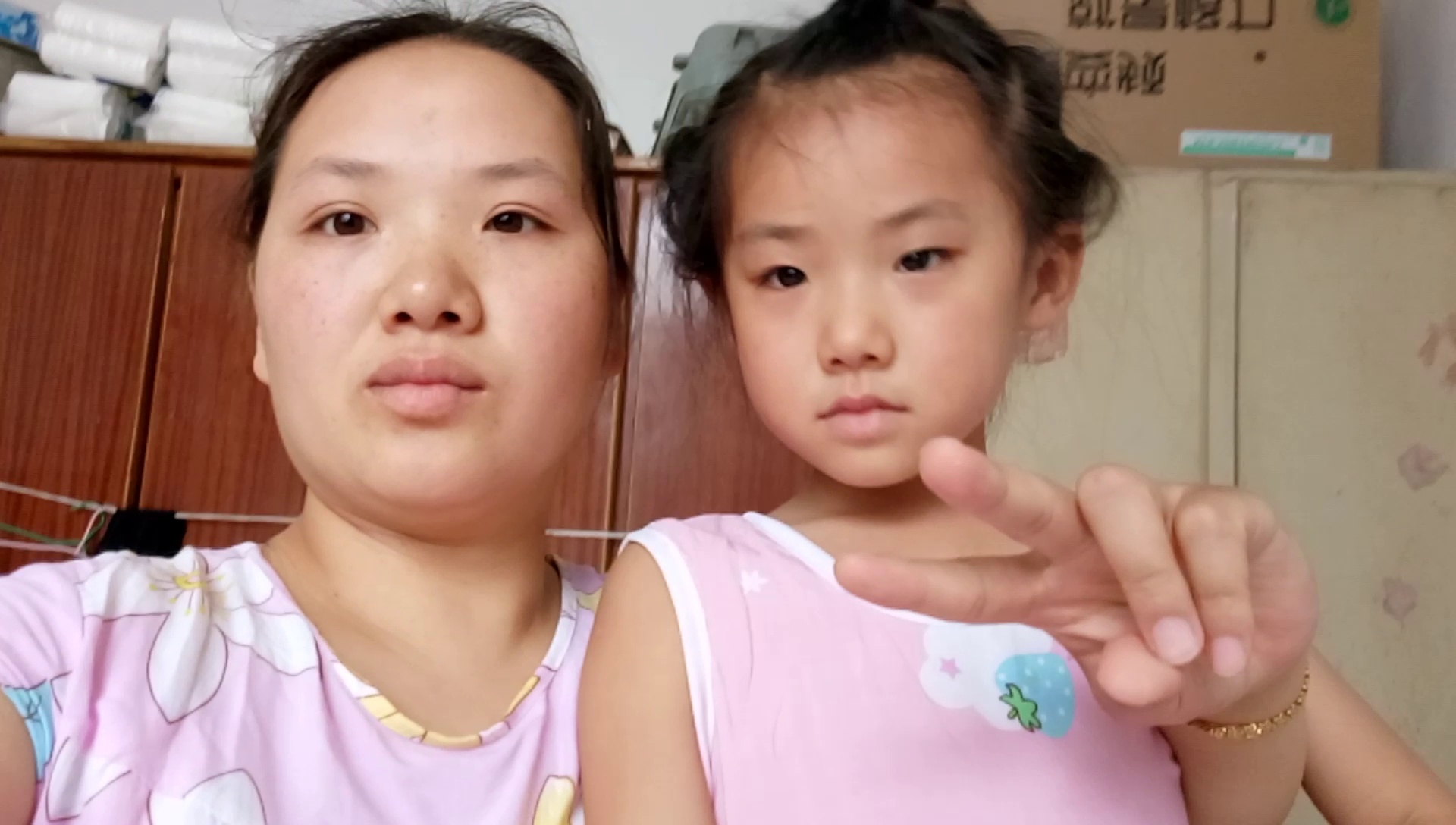} \ &
\includegraphics[width=0.15\linewidth, height=0.15\linewidth]
{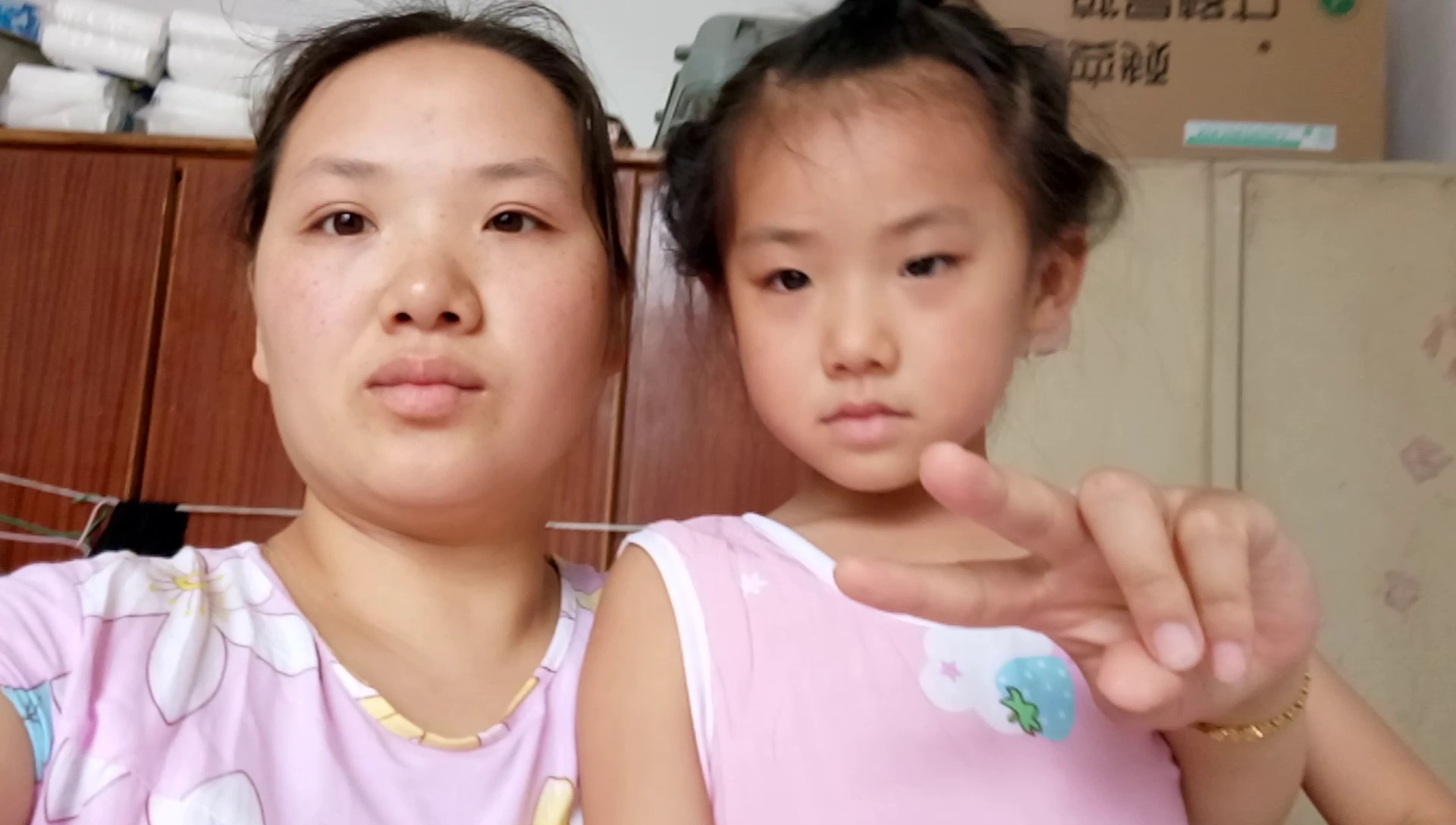} \ &
\includegraphics[width=0.15\linewidth, height=0.15\linewidth]
{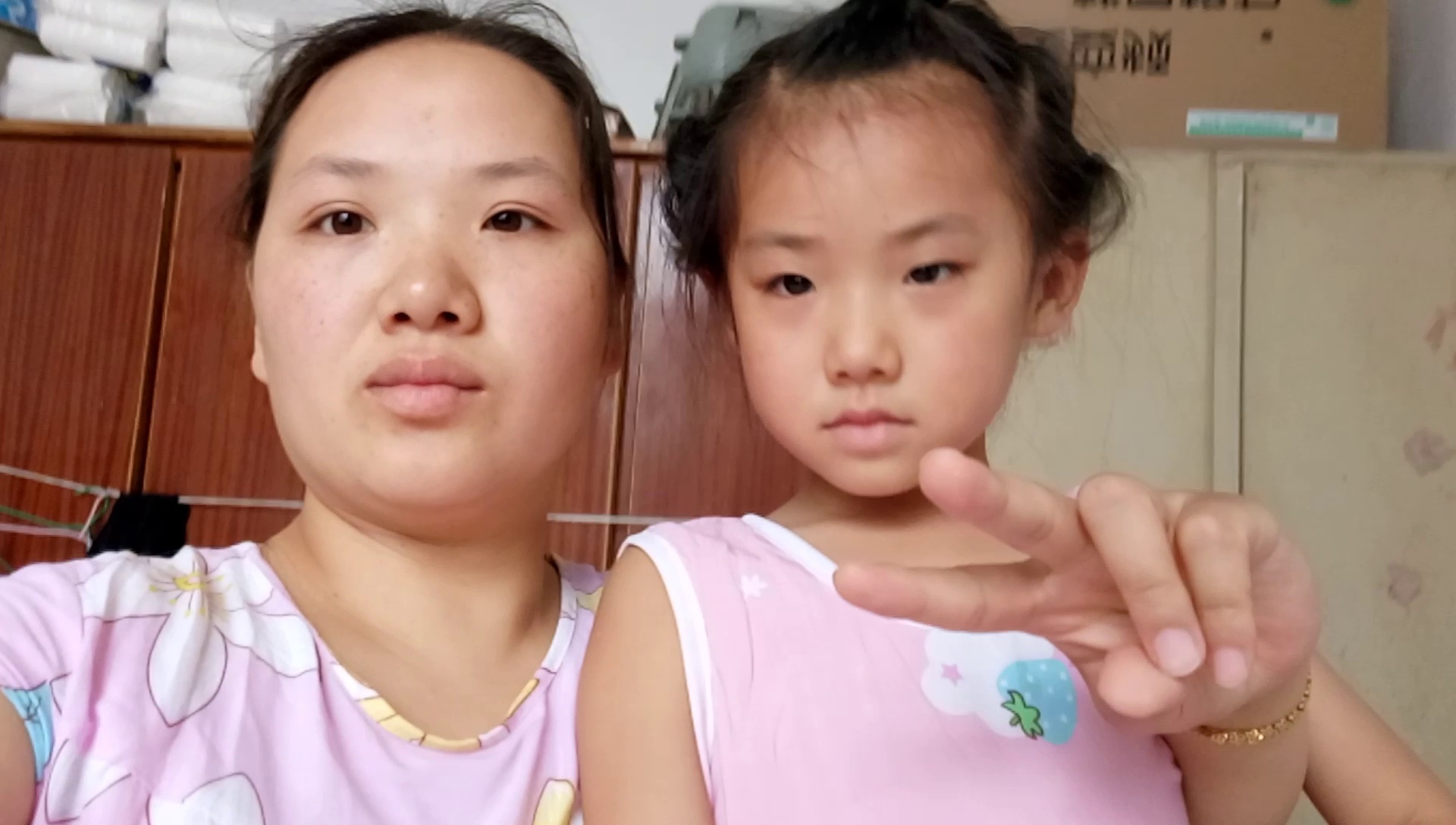} \ &
\includegraphics[width=0.15\linewidth, height=0.15\linewidth]
{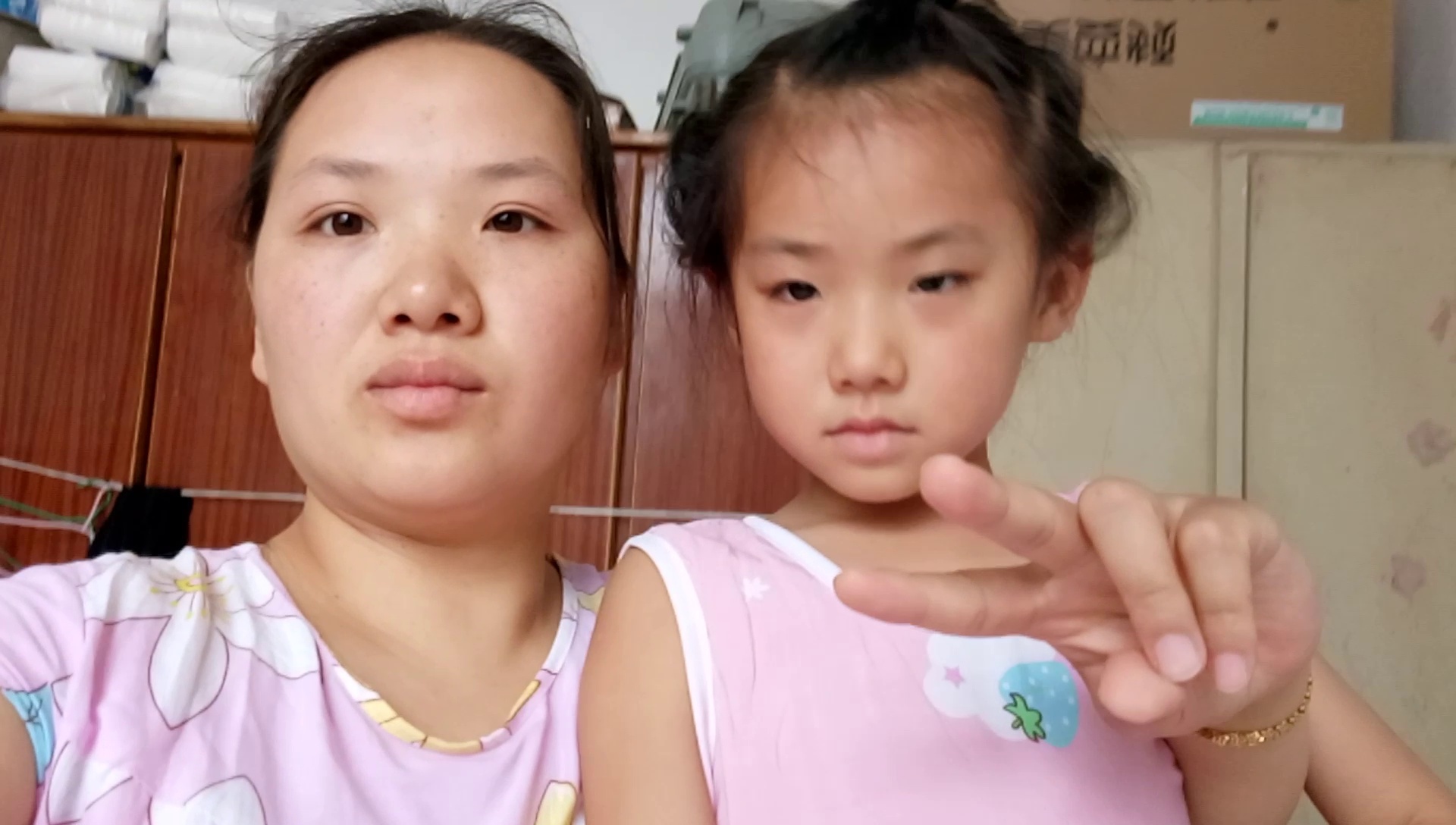} \ &
\includegraphics[width=0.15\linewidth, height=0.15\linewidth]
{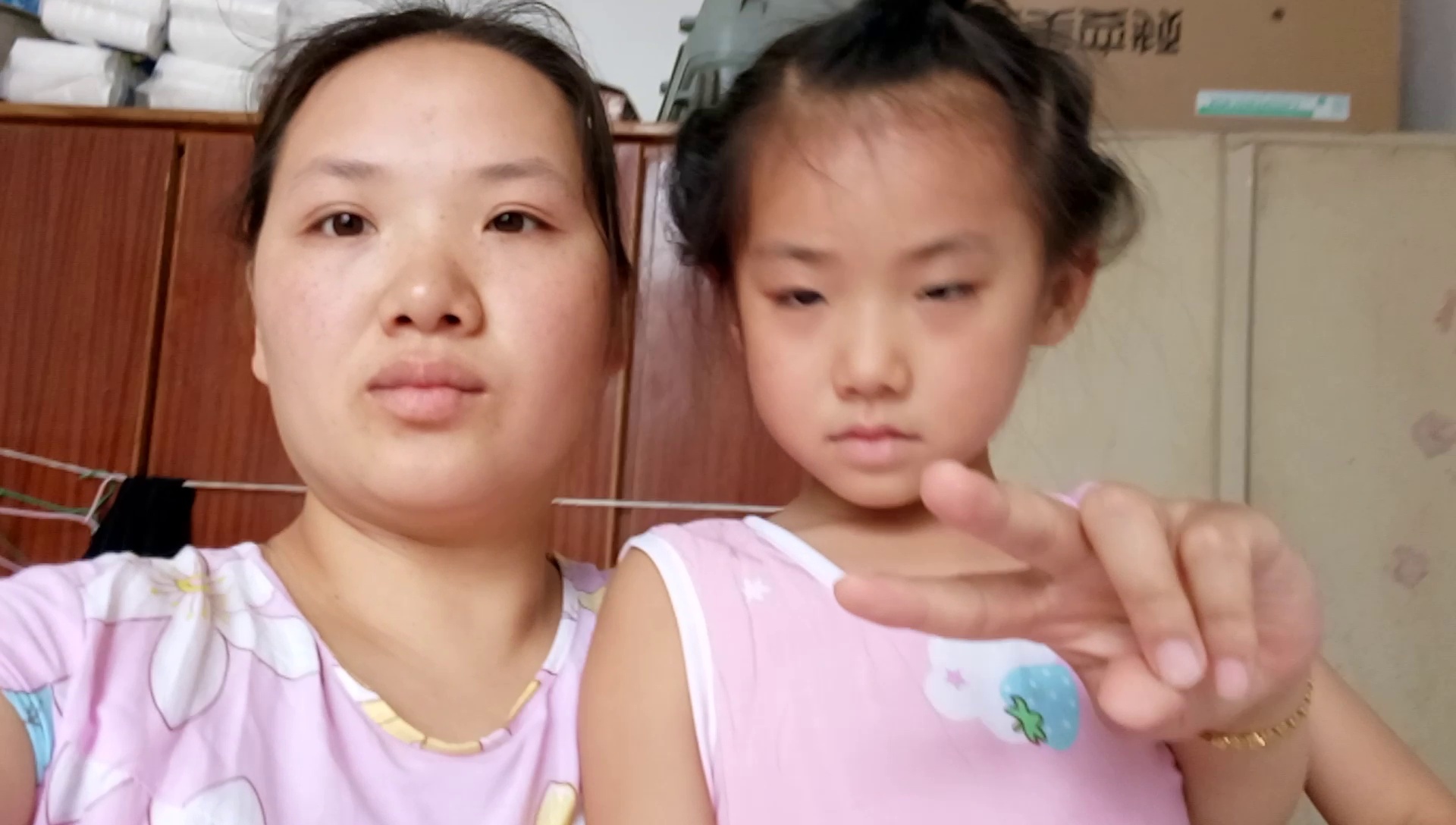}  \\

\includegraphics[width=0.15\linewidth, height=0.15\linewidth]
{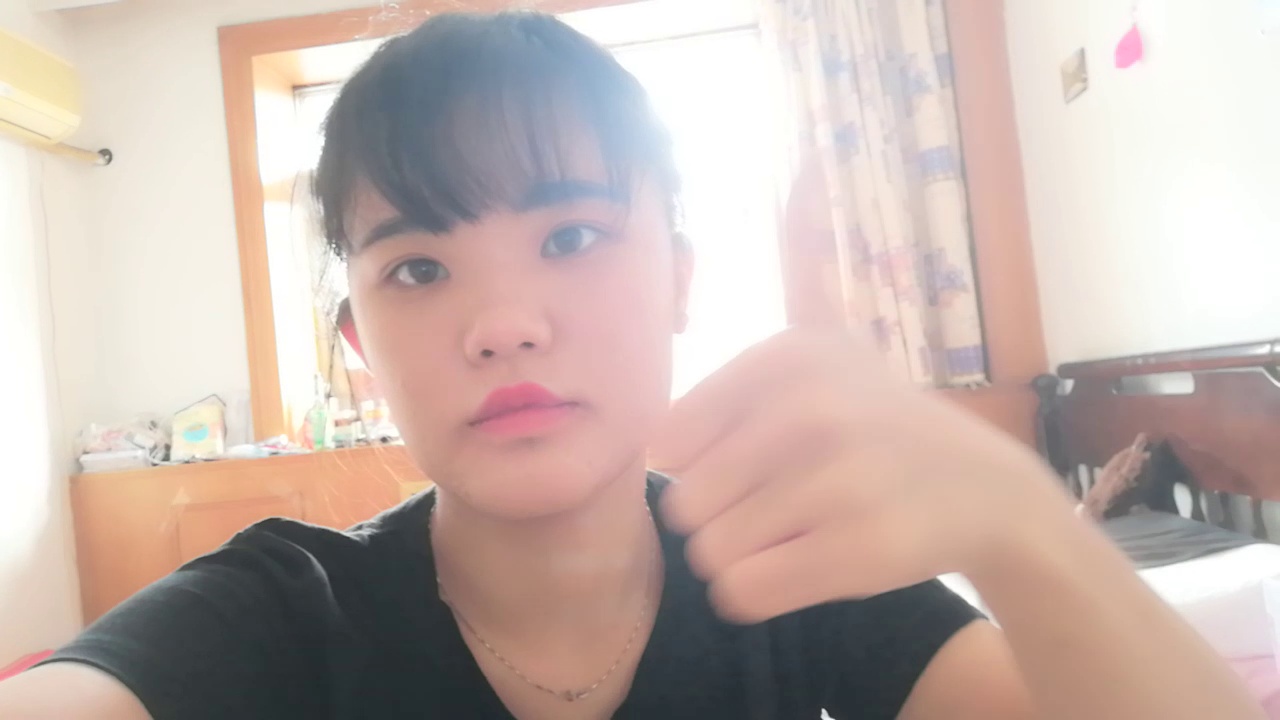} \ &
\includegraphics[width=0.15\linewidth, height=0.15\linewidth]
{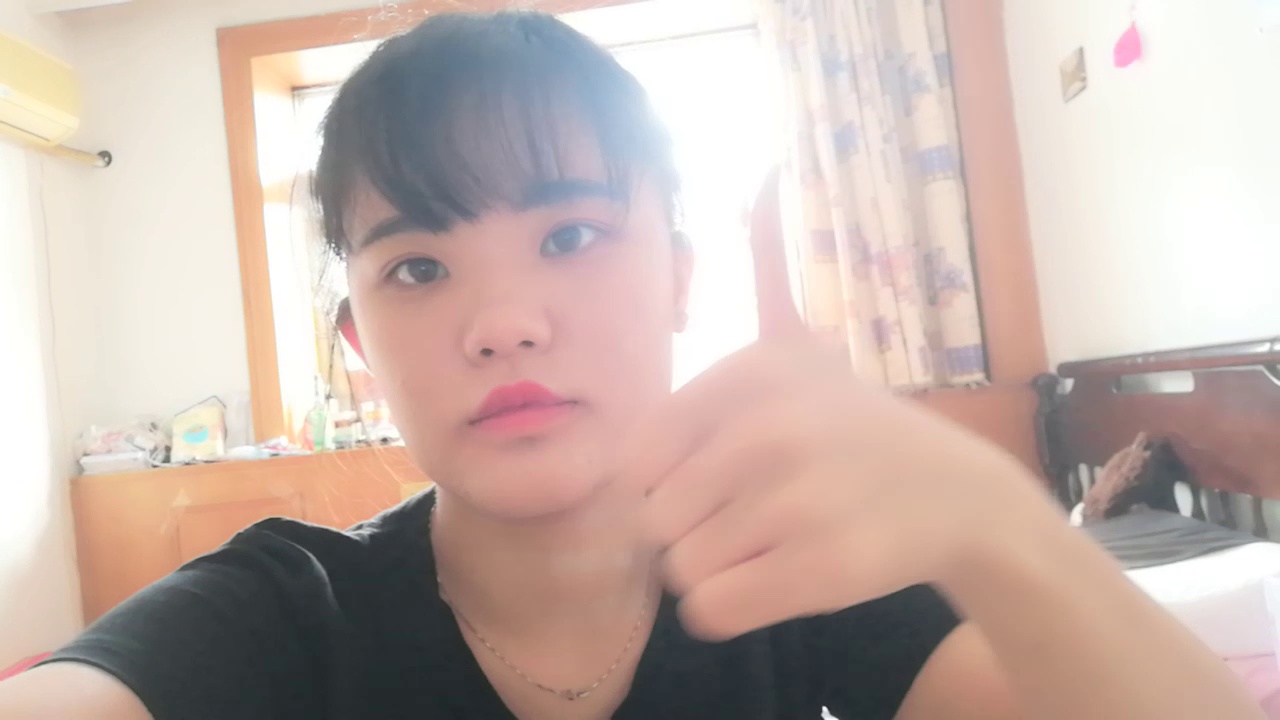} \ &
\includegraphics[width=0.15\linewidth, height=0.15\linewidth]
{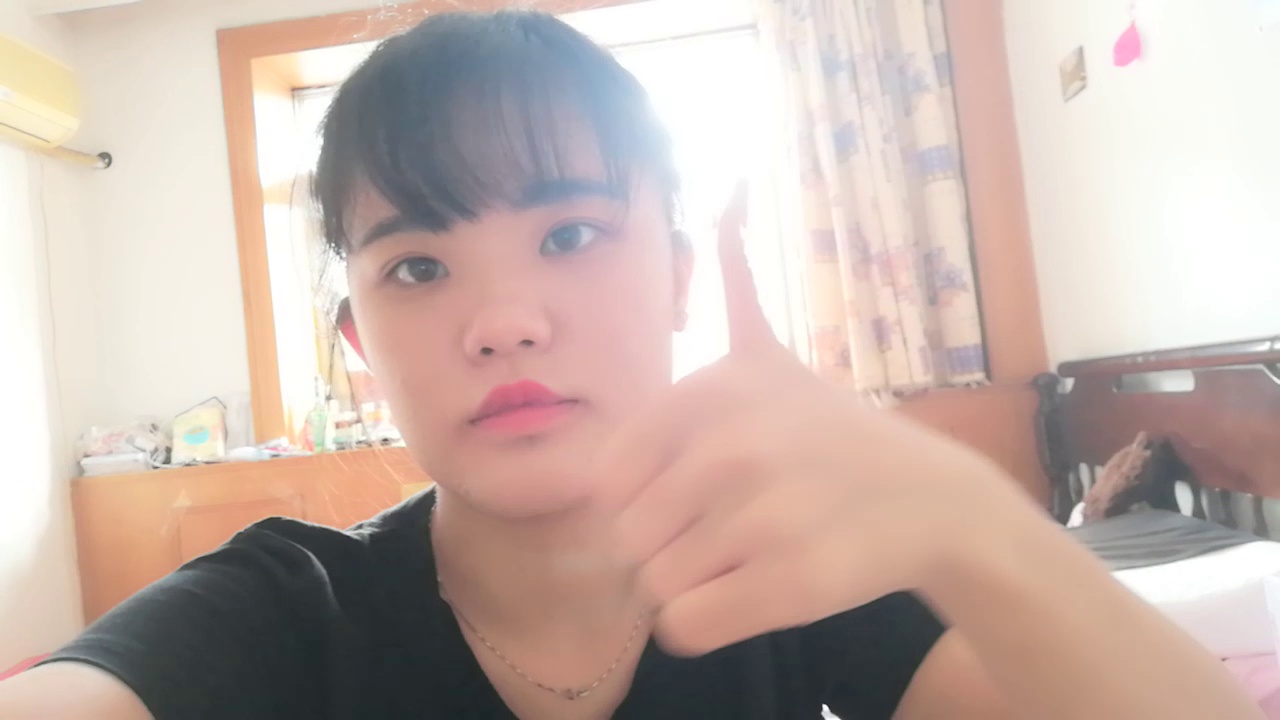} \ &
\includegraphics[width=0.15\linewidth, height=0.15\linewidth]
{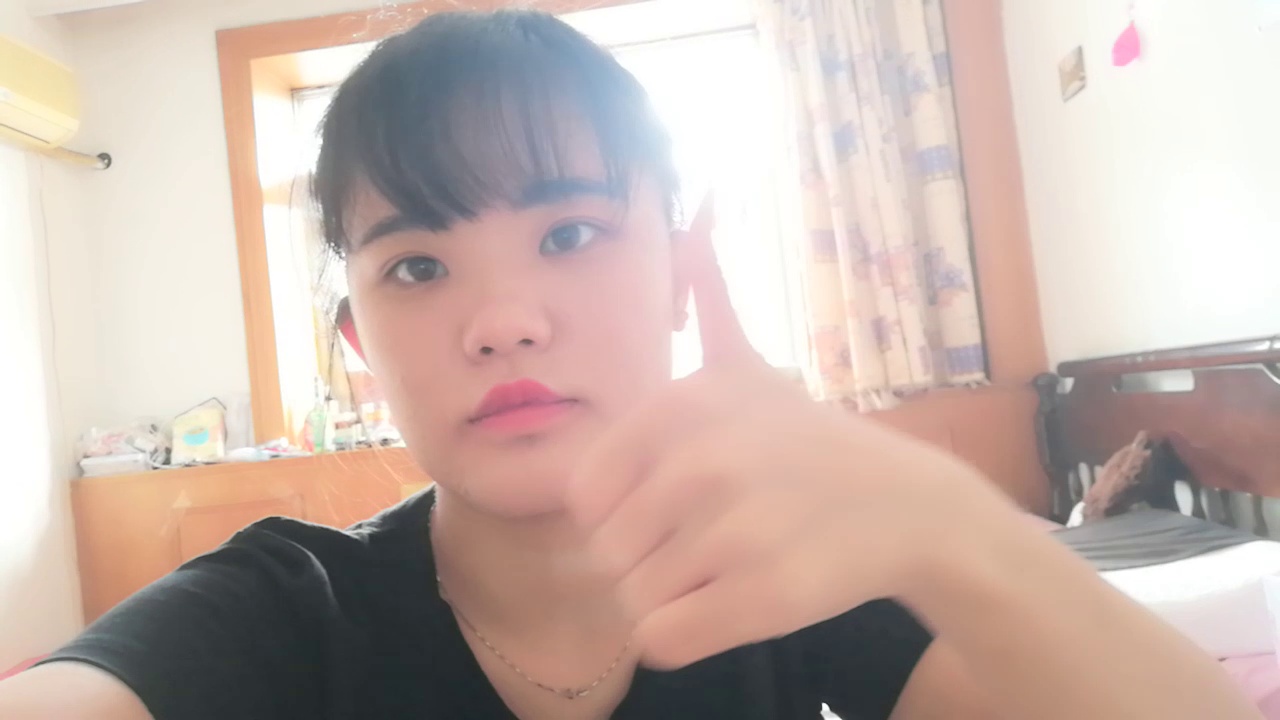} \ &
\includegraphics[width=0.15\linewidth, height=0.15\linewidth]
{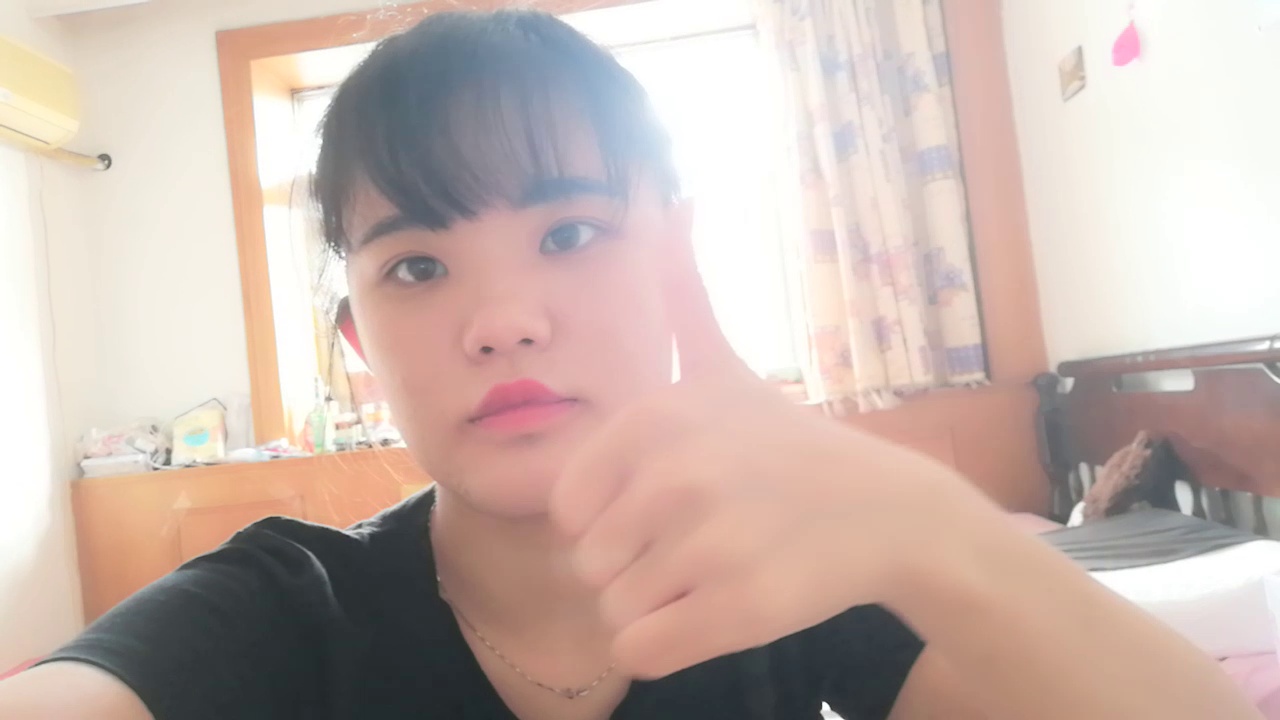} \ &
\includegraphics[width=0.15\linewidth, height=0.15\linewidth]
{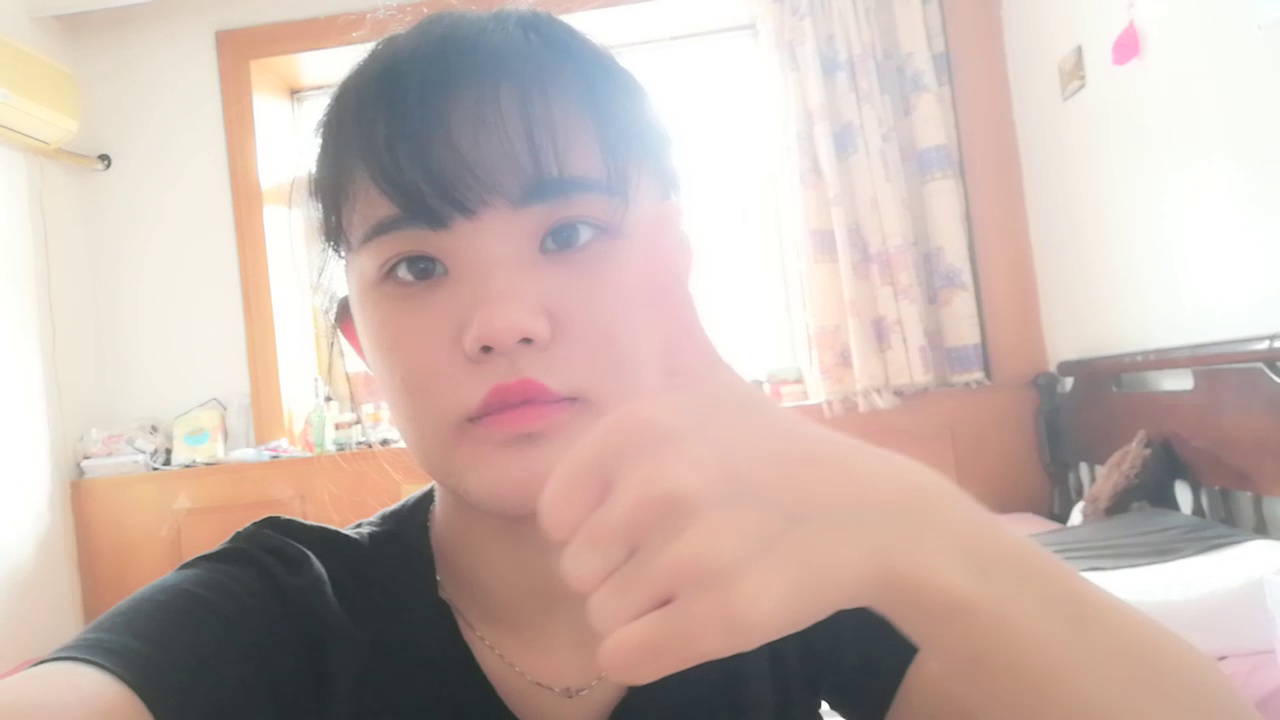}  \\
 frame $t$ & frame $t+1$ &frame $t+2$ &frame $t+3$ &frame $t+4$ & frame $t+5$
\end{tabular}
\end{center}
\caption{Videos on the video human segmentation.}
\label{Fig_results}
\end{figure*}

The labeled images in the proposed video human segmentation dataset contain about 5 parts.
(1) The images in ATR part are a total of 17,706 images of the full-body of a single person.
(2) The images in AISegment part are a total of 34,426 images of the half-body of a single person.
(3) The images in AIC part contain 3,8781 pose pictures of single to multiple people under different scenarios.
(4) The images in LIP part are a total of 5,168 images of multi people.
(5) The images in Supervisely Person Dataset part are 5,200 high quality media images under single scenes.

As there is no labels in AIC part, we invited 30 students to annotate the images and guaranteed that the error of the boundary does not exceed 3 pixels under the image size of 512.
The AISegment part was originally rough matting annotations, and we processed all the labels larger than 0.1 * 255 in the matting annotations into the foreground and the others into the background, enabling a binary annotation.
The images in LIP part were all multi people because the original labels marked the details of the neck, watch, etc. as the background in images of single person, thus we manually selected 5,168 pictures of the neck and other parts that were not marked as the background, and processed all labels of the body parts into the foreground.
The processing for images in ATR was similar to LIP.
The images in Supervisely Person Dataset(SPD) are instance-level human body annotations and we processed them into binary labels.
These 5 image parts make up about 100,000 labeled images to ensure a rich segmentation scenes of different posture, background, half-body, full-body of single or multi people.

The videos were collected by a public team of more than 100 people, and all of then were taken by using phone camera and each must contained no less than 6 actions with different camera perspectives, in any life scene containing about 1 to 5 individuals.
Thus, the video part contains 130 videos lasting from 30 seconds to 1 minute, and all the videos were sampled and got about 300 to 600 frames for each video.

\section{Results on videos}
Intuitive video performance can be found at \url{https://github.com/Ximoi/Coherent_VOS}

\begin{table*}[tbp]
\setlength{\abovecaptionskip}{0mm}
\setlength{\belowcaptionskip}{0mm}
\begin{center}
\caption{More comparison results on DAVIS set.}
\label{table:headings}
\begin{tabular}{ccccc}
\hline
Method  & mIoU(val) & MAE(val) & mIoU(val) & MAE(val) \\
\hline
        & offline  & offline  &  online   & online    \\
CRN     &   73.38  & 2.11     &  82.32    & 0.803   \\
CRN+CL  &   75.01  & 1.59     &  83.03    & 0.802   \\
SCHP    &   70.66  & 2.89     &  78.99    & 1.97      \\
SCHP+CL &   75.51  & 2.33     &  81.59    & 1.77      \\
\hline
\hline
Method  &${STB}_{global}$ &${STB}_{local}$  &${STB}_{global}$   &${STB}_{local}$\\
\hline
        & offline  &offline             & online    & online \\
CRN     & 99.07     & 93.87             & 99.45     & 95.21  \\
CRN+CL  & 99.21     & 94.52             & 99.27     & 95.35  \\
SCHP    & 98.31     & 93.42             & 99.57     & 95.37  \\
SCHP+CL & 98.79     & 94.25             & 99.46     & 95.73  \\
\hline
\end{tabular}
\end{center}
\end{table*}

\begin{table*}
\setlength{\abovecaptionskip}{0mm}
\setlength{\belowcaptionskip}{-0.1mm}
\begin{center}
\caption{More comparison on Cityscape val set. mIoU/mAcc/aAcc stands for mean IoU, mean accuracy of each class and all pixel accuracy respectively. ss denotes single scale testing.}
\label{table:pspnet}
\begin{tabular}{ccccccc}
\hline
 Method     & mIoU(val) & mAcc(val) & aAcc(ss,val)  & ${STB}_{global}$   &${STB}_{local}$ \\
\hline
PSPNet      &  77.02    &  84.17   &  95.90         &   93.06            &    90.22       \\
PSPNet+CL   &  77.88    &  85.43   &  96.04         &   93.08            &    91.63       \\
\hline
\end{tabular}
\end{center}
\end{table*}

\end{document}